\documentclass[10pt,twocolumn,letterpaper]{article}

\usepackage[pagenumbers]{cvpr} 

\usepackage[dvipsnames]{xcolor}

\usepackage{booktabs}
\usepackage{makecell}
\usepackage{multirow}
\usepackage{diagbox}
\usepackage{rotating}
\usepackage[normalem]{ulem}
\useunder{\uline}{\ul}{}
\usepackage{csquotes}

\definecolor{cvprblue}{rgb}{0.21,0.49,0.74}
\usepackage[pagebackref,breaklinks,colorlinks,citecolor=cvprblue]
{hyperref}

\usepackage{lipsum}

\def\paperID{16622} 
\def\confName{CVPR}
\def\confYear{2024}

\title{\name{}: \\ A generative language annotation framework to reveal visual biases in datasets}

\author{Krish Kabra\\
	Rice University\\
	{\tt\small kk80@rice.edu}
	\and
	Kathleen M. Lewis\\
	Massachusetts Institute of Technology\\
	{\tt\small kmlewis@mit.edu}
	\and
	Guha Balakrishnan\\
	Rice University\\
	{\tt\small guha@rice.edu}
}
\newcommand{\name}{GELDA}

\begin{document}
\maketitle

\begin{abstract}
Bias analysis is a crucial step in the process of creating fair datasets for training and evaluating computer vision models. The bottleneck in dataset analysis is annotation, which typically requires: (1) specifying a list of attributes relevant to the dataset domain, and (2) classifying each image-attribute pair. While the second step has made rapid progress in automation, the first has remained human-centered, requiring an experimenter to compile lists of in-domain attributes. However, an experimenter may have limited foresight leading to annotation ``blind spots," which in turn can lead to flawed downstream dataset analyses. To combat this, we propose \name{}, a nearly automatic framework that leverages large generative language models (LLMs) to propose and label various attributes for a domain. \name{} takes a user-defined domain caption (e.g., ``a photo of a bird," ``a photo of a living room") and uses an LLM to hierarchically generate attributes. In addition, \name{} uses the LLM to decide which of a set of vision-language models (VLMs) to use to classify each attribute in images. Results on real datasets show that \name{} can generate accurate and diverse visual attribute suggestions, and uncover biases such as confounding between class labels and background features. Results on synthetic datasets demonstrate that \name{} can be used to evaluate the biases of text-to-image diffusion models and generative adversarial networks. Overall, we show that while \name{} is not accurate enough to replace human annotators, it can serve as a complementary tool to help humans analyze datasets in a cheap, low-effort, and flexible manner.

\end{abstract}    
\section{Introduction}
\label{sec:intro}




Dataset bias analysis is a crucial step in the machine learning model development process and typically occurs when certain attribute combinations are over- or under-represented. Dataset bias virtually always exists in observational data sampled ``from-the-wild.'' For example, the popular face dataset CelebA has a low percentage of dark-skinned faces, and a significantly higher fraction of young women compared to young men~\cite{balakrishnan2021towards}, ImageNet is known to have inequalities of visual concepts across its 1000 classes~\cite{yang2020towards}, and public chest radiograph datasets are surprisingly predictive of race~\cite{gichoya2022ai}. Measuring dataset bias is the first step towards mitigating bias, which is important for two reasons. First, machine learning models can inherit biases from training data~\cite{karkkainen2021fairface,ponce2006dataset,zemel2013learning}, resulting in potentially unfair behavior when deployed. Second, evaluation data with spurious correlations between visual attributes (e.g., age with gender in CelebA) prevent experimenters from causally linking model performance to specific visual phenomena~\cite{balakrishnan2021towards}.

Sampling biases may be measured by annotating each image in a dataset with a list of labels, and computing frequency statistics over these labels. The typical annotation workflow involves two steps: (1) compiling a list of attributes to annotate, and (2) annotating those attributes for each example. In the first step, a human dataset designer/engineer typically decides upon a set of key attributes keeping in mind specific downstream use cases of the dataset. For example, face images may be labeled with various attributes that are key for face analysis systems, such as facial expression, skin tone, or perceived age. In the second step, the designer may employ crowdsourced annotators or automated algorithms to annotate the presence/absence of each attribute in each image. 

While the second step (annotation) is clearly moving rapidly towards automation with the various advances in object recognition and foundation models~\cite{kirillov2023segany, li2021align, yuan2021florence}, the first step (attribute selection) remains largely human-centered. This raises a subtle issue: the process is only as good as the attributes decided upon by human experimenters, which can leave attribute \emph{blind spots} that they might not even foresee. For example, while species, color, and size may be annotated for birds in CUB-200~\cite{wah2011caltech}, backgrounds and perching behavior are not, though they clearly have imbalances (as we show in our experiments, see Table~\ref{tab:table-bias} and Fig.~\ref{fig:background-bias}). While there is no substitute for human ground truth, an annotation method that trades off accuracy for flexibility and automation would enable practitioners to quickly and effortlessly gather insights about their dataset. We propose such a method. 



The key insight behind our method, called \name{} (for \textbf{GE}nerative \textbf{L}anguage-based \textbf{D}ataset \textbf{A}nnotation), is that generative large language models (LLMs) like GPT~\cite{brown2020language, openai2023gpt4} capture a significant amount of world knowledge~\cite{petroni2019language} and can serve as priors~\cite{yang2023language} for linking domains to their related attributes. In addition, recent work has demonstrated the effectiveness of using LLMs to select downstream models for given tasks~\cite{gupta2023visual}. Therefore, we posit that LLMs may be used to automatically curate a rich set of relevant, domain-specific attributes and select vision models suited to the ``type'' of each attribute (for example, attributes related to objects are suited for object detectors, whereas holistic image attributes, like ``color scheme" or ``style", are suited for image-text matching models).  


Provided a user-specified domain, \name{} queries an LLM (GPT in our experiments) for semantic categories (e.g., living room furniture and color scheme) and attributes per category (e.g., couch and coffee table for the furniture category) that can visually distinguish images from that domain. Second, we use vision-language models (VLMs) to annotate the generated attributes for the images conditioned on the attribute labels. We use a zero-shot captioning model to annotate attributes related to image-level concepts (e.g., background setting, style), and a text-guided object grounding algorithm to annotate attributes related to object-level concepts (e.g., object and part detection). \name{} is automatic with the exception of a few low-cost user inputs (e.g., domain caption, number of desired categories/attributes). 

We evaluate our work with both popular computer vision datasets and synthetic image data produced by state-of-the-art text-to-image models (Stable Diffusion~\cite{podell2023sdxl}) and generative adversarial networks (StyleGAN2~\cite{karras2020analyzing}). First, we demonstrate that GPT is capable of recovering a high percentage of labels already annotated in several vision datasets, while also suggesting other relevant concepts. Second, we demonstrate that \name{} can discover previously known and unknown biases in real datasets. Examples include waterbird species in CUB-200~\cite{wah2011caltech} appearing more often in ``coastal'' or ``wetland'' backgrounds than land habitats, and luxury brands in Stanford Cars~\cite{krause20133d} appearing less often in ``parking lots'' or ``gas stations'' compared to other brands. Third, we use \name{} to show that living rooms generated by Stable Diffusion almost always have neutral or monochromatic color schemes and contain coffee tables, sofas, area rugs, and throw pillows, and that StyleGAN2 amplifies biases from its training set. Finally, we present some of \name{}'s limitations and draw conclusions regarding the safe use of this new data analysis framework. 

%


\begin{figure*}[t!]
    \includegraphics[width=\textwidth]{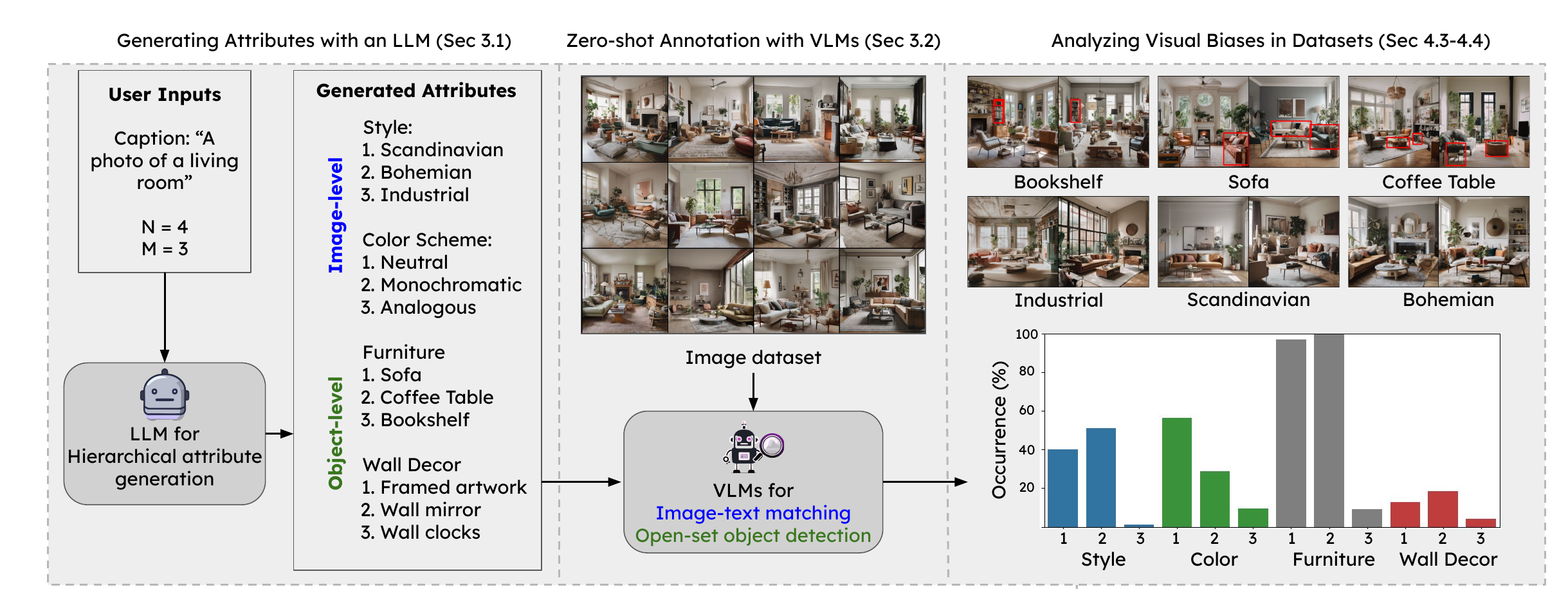}
   \caption{\textbf{Overview of \name{}.} Given a user-specified domain in the form of a caption, \name{} first queries an LLM to generate a set of visual attributes to annotate for an image dataset from that domain. The querying method is hierarchical, in that \name{} prompts the LLM to first generate \textit{N} attribute categories, then generate $M$ labels per attribute category, and finally describe whether each attribute is object-level or image-level. In the second stage, \name{} uses pre-trained VLMs to automatically annotate the generated attributes for each image. We use the LLM to assign all image-level attributes to a VLM tuned for image-text matching, and all object-level attributes to a VLM for open-vocabulary object detection. Once GELDA has identified and annotated visual attributes, we can then analyze visual biases in the dataset.}
   \label{fig:pipeline}
\end{figure*}

\section{Related Works}
\label{sec:related-works}

\subsection{Dataset bias measurement in computer vision} 
Computer vision datasets are known to have biases~\cite{ponce2006dataset,tommasi2017deeper,torralba2011unbiased,wang2021directional,wang2022revise,yang2020towards}, and human-related domains such as faces are particularly scrutinized~\cite{albiero2020analysis,drozdowski2020demographic,klare2012face,kortylewski2018empirically,kortylewski2019analyzing,merler2019diversity} because models trained on these data can inherit biases along attributes like race and gender that are protected by the law~\cite{kleinberg2019,ramaswamy2021fair,zietlow2022leveling}. Biased benchmarking data can inhibit causal analysis of algorithmic performance due to specific visual factors, due to confounding variables~\cite{balakrishnan2021towards,liang2023benchmarking}. Approaches to mitigating dataset bias include collecting more thorough examples~\cite{merler2019diversity}, using image synthesis to fill distribution gaps~\cite{kortylewski2019analyzing,ramaswamy2021fair}, and resampling~\cite{li2019repair}.

Our work is most closely related to and inspired by REVISE~\cite{wang2022revise}, a recent dataset bias analysis tool that also computes visual attribute frequencies. The main distinction between REVISE and \name{} is that REVISE relies on ground truth dataset annotations and focuses on three axes of analysis (object, person, and geography) on natural scenes, while \name{} uses LLM/VLMs to generate and label attributes, and is best suited for closed domain datasets. As a consequence, \name{} sacrifices some accuracy for flexibility and automation. 

Several other tools have also been developed to diagnose the weaknesses of machine learning models such as object detectors and action recognizers~\cite{alwassel2018diagnosing,hoiem2012diagnosing,sigurdsson2017actions}. Facebook’s Fairness Flow~\cite{facebook} and IBM's AI Fairness 360~\cite{bellamy2018ai} focus on assessing machine learning model biases as opposed to dataset biases. Amazon SageMaker Clarify~\cite{clarify} also works to detect bias in training data, but along predefined axes. Google’s Know Your Data~\cite{google} also aims to help mitigate bias issues in image datasets, but their tool currently only works on TensorFlow image datasets. In contrast, \name{} uses LLMs and VLMs to automatically generate annotations for any dataset from a specific domain.

\subsection{Foundation models and zero-shot learning} 
Foundation models are large-scale machine learning models pre-trained on vast amounts of data to learn general patterns~\cite{Bommasani2021FoundationModels, yuan2021florence}. These models serve as fundamental building blocks for various AI applications. Large language models (LLMs), like GPT-3~\cite{brown2020language} and its successors~\cite{openai2023gpt4}, have made significant advancements in natural language understanding and generation. They are widely used in various applications, including chatbots, content generation, and language translation. Vision-language models (VLMs), trained on large-scale multimodal datasets, have been shown to perform strongly on downstream tasks such as zero-shot classification, image captioning, and object detection~\cite{radford2021learning, minderer2023scaling, li2021align, wang2022git, yu2022coca}. Recent works demonstrate the power of combining multiple foundation models to perform tasks. Several works have shown improved classification performance by first prompting LLMs to generate class-specific text descriptions and then using a VLM to combine images and the generated text for classification~\cite{yang2023language,zhang2023prompt, maniparambil2023enhancing,lewis2023gist}. Another work shows that LLMs can be used to augment text in image-text datasets to assist in zero-shot classification~\cite{fan2023improving}. GELDA uses a foundation model composition, but for a new purpose: identifying and labeling domain-specific attributes in image datasets using LLMs and VLMs for bias analysis.
\section{Methods}
\label{sec:methods}
Our goal is to take a user-specified domain along with a set of images $S$ from that domain, and automatically produce attribute annotations for each image in $S$ from a variety of in-domain categories. Using these attributes, we can then perform bias analyses of $S$. There are two key challenges to this task: (1) automatically obtaining a list of relevant categories and attributes for the specified domain, and (2) automatically choosing the appropriate model for evaluating each image-attribute pair. We propose a framework (see~\cref{fig:pipeline}) that addresses both of these challenges. 

Our insight for the first challenge is that large language models (LLMs) are adept at linking concepts to one another~\cite{petroni2019language, yang2023language}. We therefore query an LLM for a list of domain categories along with their associated attributes with careful prompting. To address the second challenge, we observe that vision-language models (VLMs) offer a powerful means of performing such evaluations like zero-shot image classification~\cite{radford2021learning} and object grounding~\cite{minderer2023scaling} from text input alone. The key challenge is determining which VLM to use for a given attribute. Certain image-level attributes like style or color scheme are better suited for image-text matching (ITM) models, whereas determining the presence of an object like a couch is better suited for open-vocabulary object detectors (OVODs). We again use the LLM, this time to provide a decision into the attribute type, and automatically choose the appropriate VLM based on a pre-specified list of VLMs for each attribute type. We describe our method further in the following sections.

\subsection{Attribute generation with an LLM}
\label{subsec:using-llm-to-gen-attrs}
We use an LLM to generate attributes in a hierarchical fashion by querying the LLM for categories, followed by querying attribute examples per category. We use this hierarchical form for several reasons. First, we empirically find that querying the LLM directly for attributes results in poor coverage of visual concepts. Second, breaking up the prediction as a ``chain'' is known to be a successful strategy for controlling LLMs towards more human-like reasoning~\cite{wei2022chain}. Third, this approach allows the user control over the number of categories and attributes per category that they desire. First, the user provides a prompt query $Q1$ of the form:
\vspace{-0.15cm}
\begin{displayquote}
$Q1:$ ``What are $N$ attribute categories that can be used to visually distinguish images described by the caption \texttt{caption}?'',
\end{displayquote}
\vspace{-0.15cm}
where $N$ is a number chosen by the user and \texttt{caption} is a word or phrase describing the data domain (e.g., ``birds'' or ``a headshot photo of a person''). Second, for each of the categories \texttt{\{category1, \dots, categoryN\}} returned by $Q1$, we obtain attribute labels with query $Q2$:
\vspace{-0.15cm}
\begin{displayquote}
$Q2:$ ``What are $M$ different examples of the category \texttt{category} that can be used to distinguish images described by the caption \texttt{caption}?'',
\end{displayquote}
\vspace{-0.15cm}
where $M$ is again chosen by the user. Lastly, we determine whether each of the $N$ attribute categories relates to image-level or object-level concepts with query $Q3$:
\vspace{-0.15cm}
\begin{displayquote}
$Q3$: ``Are \texttt{\{att1, \dots, attM\}} examples of objects or items? Answer with a yes or no. Explain your answer.'', 
\end{displayquote}
\vspace{-0.15cm}
where \texttt{\{att1, att2, \dots attM\}} is the list of $M$ generated attributes for a category. We require a binary yes or no answer in order to automatically filter the response into one of the two appropriate downstream models. Requiring an explanation pushes the model to provide more accurate answers, as demonstrated in prior work~\cite{wei2022chain}.

\textbf{Dealing with stochasticity:} Auto-regressive LLMs are stochastic in that they can produce different outputs given the same prompt. While stochasticity helps capture the full output distribution, determinism is helpful for reproducibility. To obtain high-quality attribute labels that are mostly consistent across experiments, we perform the queries in the previous section several times per prompt, and pick the $N$ and $M$ most frequently labeled categories and attributes. 



\subsection{Zero-shot annotation with VLMs}
We assume access to pretrained VLMs that take input images and text captions and can perform annotation. In our experiments, we use two VLMs -- one for image-text matching (ITM) and one for open-vocabulary object detection (OVOD). To convert LLM-generated attributes into input captions for the VLMs, we define a set of prompt templates that correspond to noun-attribute relationship phrases, e.g., ``a \texttt{noun} has \texttt{attribute}'' and ``a \texttt{attribute} \texttt{noun}.'' We let the user assign the correct noun-attribute relationship phrase for each of the $N$ categories, though this can likely also be automated by the LLM in future work.

OVOD models output bounding boxes and detection scores, allowing us to label an attribute if its detection score is simply above a threshold $\alpha$. Output values of current ITM models are less predictable because they are trained with a hard negative mining strategy~\cite{li2021align}, making it difficult to set a constant threshold. Instead, we compute ITM scores for the $M$ attribute text captions and a generic ``base'' reference caption describing the domain (same as the one used in query $Q1$, see Sec.~\ref{subsec:using-llm-to-gen-attrs}). Finally, we select the highest-scoring caption among the $M$ attributes, and label that attribute as present if it is greater than the base caption score. This process essentially performs multiclass classification.



\section{Experiments and Results}
\label{sec:experiments}
We evaluate \name{} in several ways. Sec.~\ref{subsec:eval-llm} and Sec.~\ref{subsec:eval-vlms} quantitatively analyze the performances of the LLM and VLM components. Sec.~\ref{subsec:exp-real-datasets} demonstrates visual biases discovered by \name{} in real datasets, and Sec.~\ref{subsec:exp-synthetic-datasets} demonstrates biases discovered in deep generative model outputs. We use the following publicly available models: GPT-$3.5$ for chat completion, 
BLIP 
for ITM, and OWLv2 
for OVOD using a threshold of $\alpha=0.3$. 

\textbf{Datasets:} We used 7 real and synthetic datasets spanning a range of domains to show the generality of our method. We use the test sets of four popular real image datasets: \textbf{(1) DeepFashion} (clothing items)~\cite{liuLQWTcvpr16DeepFashion}, \textbf{(2) CelebA} (human faces)~\cite{liu2015faceattributes}, \textbf{(3) CUB-200} (birds)~\cite{wah2011caltech}, and \textbf{(4) Stanford Cars} (cars)~\cite{krause20133d}. We use the public Stable Diffusion XL model 
~\cite{podell2023sdxl} to generate \textbf{(5) SD Living Rooms}, consisting of 1,024 synthetic images using the caption ``a photo of a living room.'' We use the public StyleGAN2 (SG2) models~\cite{karras2020analyzing} 
on the FFHQ~\cite{karras2019style} and AFHQ~\cite{choi2020stargan} datasets to generate 10,000 images each of \textbf{(6) SG2 Faces} and \textbf{(7) SG2 Dogs} (with truncation $\psi=0.7$~\cite{brock2018large}). 

\begin{figure}[t!]
    \centering
    \includegraphics[width=\linewidth]{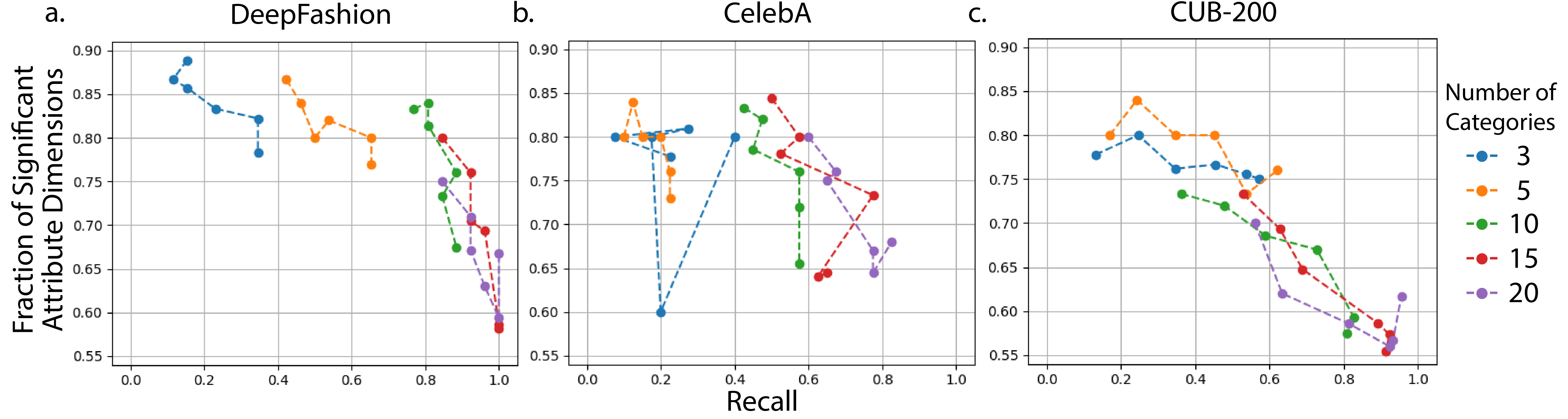}
    \caption{ \textbf{Analysis of GPT attribute generation performance on three real datasets.} We plot the fraction of significant attribute dimensions (the fraction of principal components that explain 95\% of the cumulative variance) versus recall (fraction of real dataset attributes that match with a generated attribute, see Eqn.~\ref{eqn:recall}). We plot separate curves for each number of generated categories ($N$ in Sec.~\ref{subsec:using-llm-to-gen-attrs}) queried by GPT, and each dot represents a different number of queried attributes per category ($M$).}
    \label{fig:gpt-sweep-results}
\end{figure}

\subsection{Analysis of attribute generation (LLM)}
\label{subsec:eval-llm}
GPT-3.5 virtually always generates attributes relevant (i.e., related in some manner) to a given data domain (see Supplementary for a list of generated attributes per domain). However, simply generating a huge number of attributes (large $M$ and $N$) is a poor strategy for several reasons. First, during data analysis, we want a compact set of features to avoid multi-hypothesis testing. Second, each new added attribute will eventually yield marginal information gain, leading to many redundant attributes. Third, though relatively minor, there is a cost (monetary and time) associated with each query to GPT-3.5 (\$0.0020/1K output tokens, average response time of $\sim$30 seconds/1K output tokens). 

With this in mind, we first explore the quality of the attributes generated by GPT with varying values to $M$ and $N$. We consider two figures of merit: recall and effective attribute dimension. Recall is the fraction of real labels that are annotated by GPT, measuring the ability to recover known relevant attributes. We estimate it as:
\begin{equation}
\label{eqn:recall}
    \text{Recall} = \frac{1}{|\mathcal{A_R}|} \Big( \sum_{a_r \in \mathcal{A_R}} \max_{a_g \in \mathcal{A_G}} \cos \big( \mathcal{T}(a_r), \mathcal{T}(a_g) \big) > \beta \Big),
\end{equation}
where $\mathcal{A_R}$ is the set of the real dataset attributes, $\mathcal{A_G}$ is the set of generated attributes, $\mathcal{T}(\cdot)$ is a text embedding function, $\cos(\cdot, \cdot)$ is the cosine similarity between two vectors, and $\beta$ is a threshold for defining a match in similarity. For our experiments, we set $\beta = 0.8$. We define effective attribute dimension as the number of principal components needed to capture 95\% of the variance in the embedding space spanned by a set of attributes, $\{\mathcal{T}(a_g)\}_{a_g \in A_G}$, which intuitively captures the fraction of meaningful attribute dimensions in the set. 

\begin{table}[t!]
\small
\centering
\caption{\textbf{VLM performances on ground-truth attribute labels from real datasets}. Numbers are average AUC scores across all attributes in each dataset. BLIP achieves good performance across all three datasets, and consistently outperforms OWLv2 likely due to more image-level attributes being labeled.}
\begin{tabular}{l|ccc}
\diagbox{\textit{Model}}{\textit{Dataset}} & DeepFashion & CelebA & CUB-200 \\ \hline
BLIP & 0.78 & 0.74 & 0.70 \\
OWLv2 & 0.65 & 0.71 & 0.67
\end{tabular}
\label{tab:vlm-results}
\end{table}

\begin{figure}[t!]
    \centering
    \includegraphics[width=\linewidth]{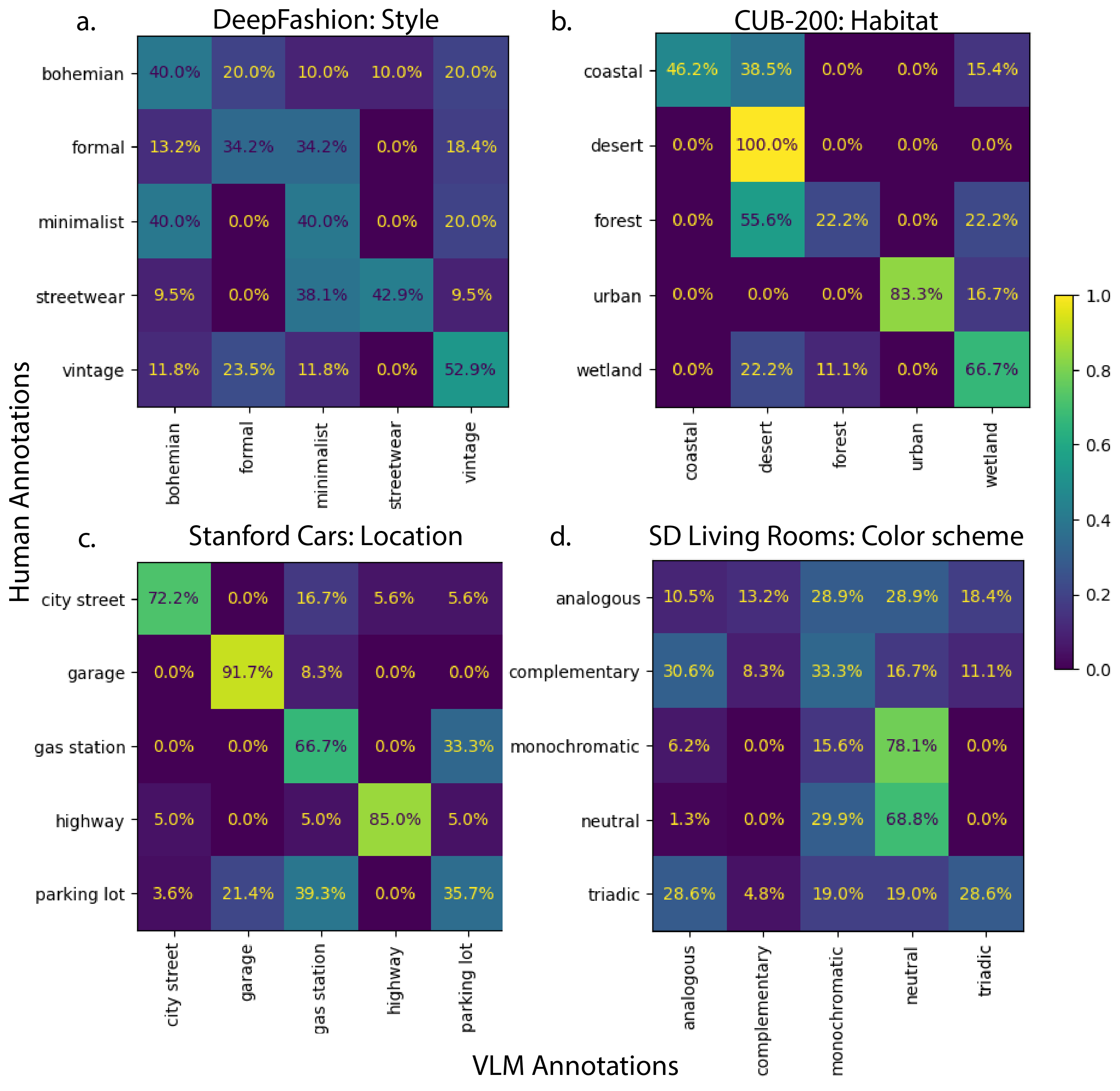}
    \caption{\textbf{Human evaluation of BLIP (VLM) for image-level annotations.} We selected one category each from four datasets for human annotation. We find that BLIP annotations match with humans for backgrounds seen in CUB-200 and Stanford Cars. However, BLIP annotations of DeepFashion style and SD Living Rooms color schemes match less strongly. Counts are normalized over rows (i.e., total human annotations) to yield percentages.}
    \label{fig:human-eval}
\end{figure}

\begin{table*}[t!]
\scriptsize
\centering
\caption{\textbf{Categories generated by \name{} along with their associated attributes with the highest frequencies for various datasets.} \uline{Category} corresponds with attributes determined to be object-level, and the majority attribute names have been shortened for brevity. Across all three datasets, several categories contain attributes with frequencies over 30\%, which may be cause for concern. More detailed breakdowns for each dataset are shown in Supplementary.}
\begin{tabular}{@{}llc@{}}
\multicolumn{3}{l}{\textbf{DeepFashion}} \\
\multicolumn{3}{l}{\textit{"a photo of a clothing item"}} \\ \hline
\textit{Category} & \textit{\begin{tabular}[c]{@{}l@{}}Majority \\ Attribute\end{tabular}} & \textit{\begin{tabular}[c]{@{}c@{}}Majority \\ Proportion (\%)\end{tabular}} \\ \hline
Material & Cotton & 23.0\% \\
Fit & Loose & 42.0\% \\
Style & Minimalist & 35.3\% \\
Pattern & Floral & 4.3\% \\
Sleeve length & Sleeveless & 30.5\% \\
Color & Black & 12.2\% \\
\uline{Embellishments} & Lace & 11.5\% \\
Neckline & Boat & 28.1\% \\
Brand/logo & Nike & 0.7\% \\
\uline{Type} & Dress & 46.2\% 
\end{tabular}
\hfil
\begin{tabular}{@{}llc@{}}
\multicolumn{3}{l}{\textbf{CUB-200}} \\
\multicolumn{3}{l}{\textit{"a photo of a bird"}} \\ \hline
\textit{Category} & \textit{\begin{tabular}[c]{@{}l@{}}Majority \\ Attribute\end{tabular}} & \textit{\begin{tabular}[c]{@{}c@{}}Majority \\ Proportion (\%)\end{tabular}} \\ \hline
Habitat & Forest & 12.5\% \\
Plumage pattern & Solid & 37.7\% \\
Color & Yellow & 11.7\% \\
Size & Tiny & 50.8\% \\
Wing shape & Slender & 10.0\% \\
Species & Sparrow & 29.5\% \\
\uline{Perching behavior} & Tree branch & 37.1\% \\
Background & Tree branch & 16.3\% \\
Shape & Circle & 1.0\% \\
Eye color & Yellow & 7.7\% 
\end{tabular}
\hfil
\begin{tabular}{@{}llc@{}}
\multicolumn{3}{l}{\textbf{Stanford Cars}} \\
\multicolumn{3}{l}{\textit{"a photo of a car"}} \\ \hline
\textit{Category} & \textit{\begin{tabular}[c]{@{}l@{}}Majority \\ Attribute\end{tabular}} & \textit{\begin{tabular}[c]{@{}c@{}}Majority \\ Proportion (\%)\end{tabular}} \\ \hline
\uline{Body type} & Coupe & 33.7\% \\
Color & Silver & 26.5\% \\
Condition & Brand new & 63.6\% \\
Year & 2010 & 43.5\% \\
Size & Mid-size & 44.9\% \\
\uline{Make/model} & Chevrolet corvette & 13.5\% \\
Lighting & Natural sunlight & 9.3\% \\
Location & Parking lot & 30.1\% \\
Features & Object detection & 0.0\% \\
Surroundings & Crowded parking lot & 8.7\% 
\end{tabular}
\label{tab:table-bias}
\end{table*}

\begin{figure}[t!]
    \centering
    \includegraphics[width=\linewidth]{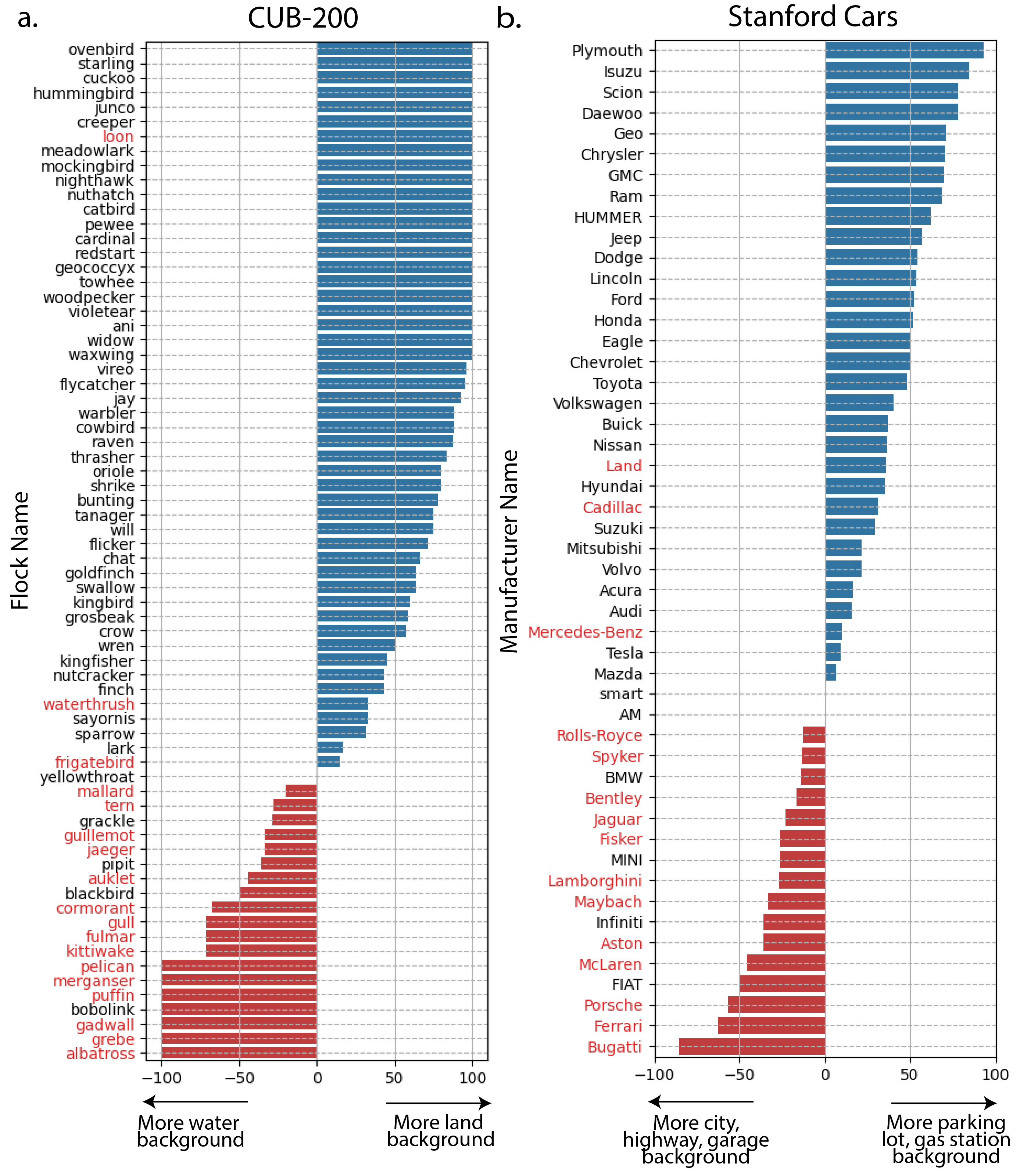}
    \caption{ \textbf{Discovered confounding relationships between class labels in CUB-200 and Stanford Cars and environmental attributes generated by \name{}.} (a.) Land versus water background bias in CUB-200. Bird species known generally as ``waterbirds'' (names in red) appear more often with water backgrounds. (b.) Location bias in Stanford Cars. Luxury brands (names in red) appear less often in ``parking lots'' or at ``gas stations" compared to ``garages''.}
    \label{fig:background-bias}
\end{figure}
\begin{figure*}[t!]
    \centering
    \includegraphics[width=1\linewidth]{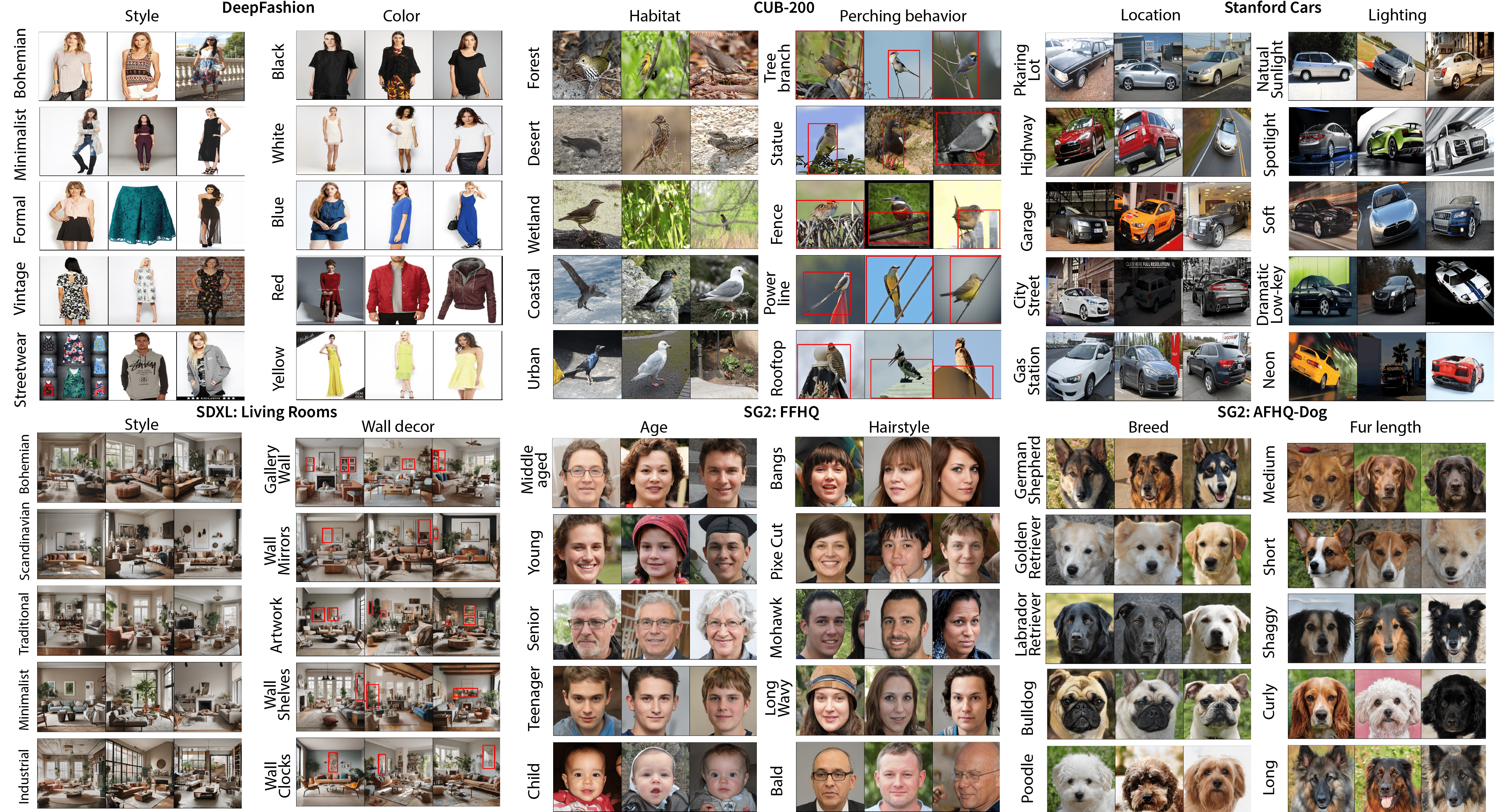}
    \caption{ \textbf{Visual samples with annotations produced \name{}.} The attribute names were generated automatically by the LLM (GPT), and the assignment of images to labels was produced by VLMs (BLIP and OWLv2). Attributes determined to be object-level are shown in images with red bounding box detections. Attribute names have been shortened for brevity.}
    \label{fig:qualitative-examples}
\end{figure*}

\begin{figure*}[t!]
    \centering
    \includegraphics[width=0.95\linewidth]{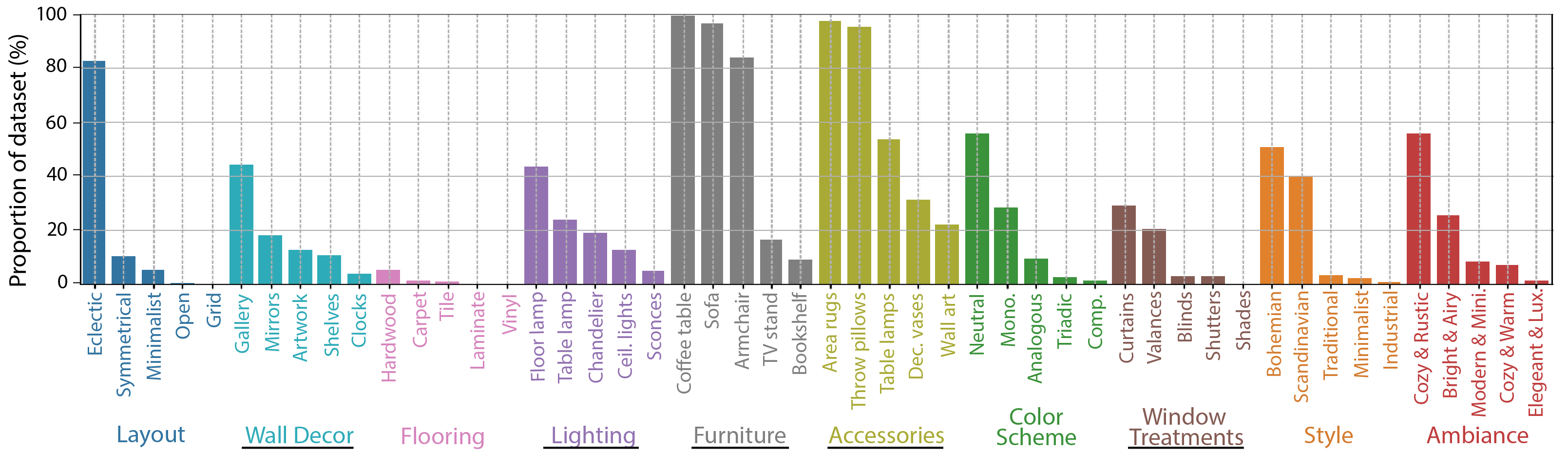}
    \caption{\textbf{Distribution of annotated attributes returned by \name{} for the SD Living Room (synthetic) dataset.} Bars are grouped by attribute categories in different colors, and attribute names have been shortened for brevity. \uline{Category} corresponds with attributes determined to be object-level. Certain attributes are prominent in the generated images, such as coffee tables and sofas for furniture, throw pillows and area rugs for accessories, neutral and monochromatic hues for color schemes, and Bohemian and Scandinavian styles. }
    \label{fig:living-rooms-dist}
\end{figure*}

\begin{figure}[t!]
    \centering
    \includegraphics[width=\linewidth]{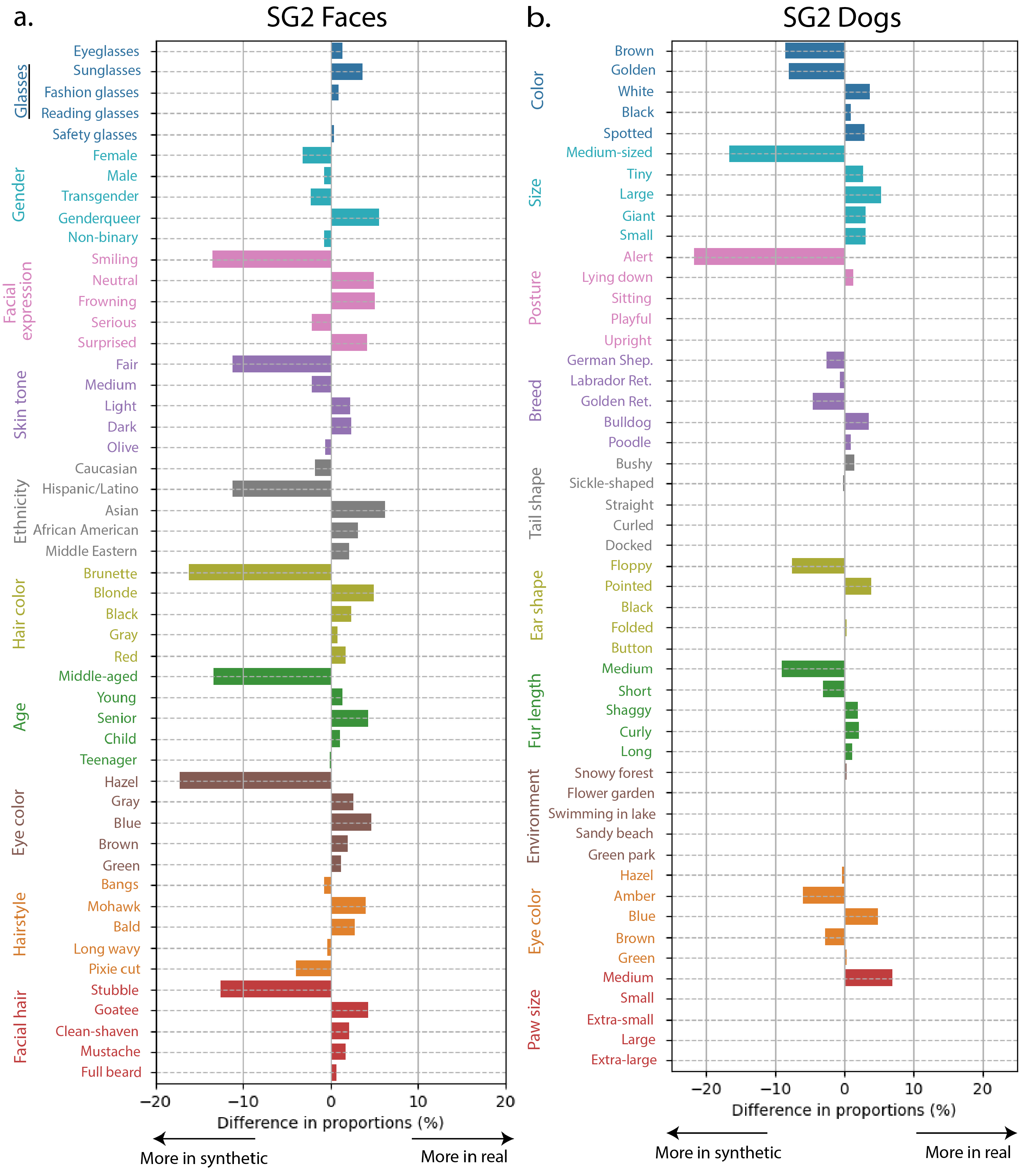}
    \caption{ \textbf{Comparisons of attribute bias of synthetic StyleGAN2 (SG2) image generators with respect to their training distributions.} (a.) SG2 Faces vs. FFHQ. (b.) SG2 Dogs vs. AFHQ. The attributes are ordered from top to bottom in each category by descending frequency in the training dataset. SG2 amplifies bias -- the most popular attributes in the training dataset for each category have an even greater majority in the generated dataset, as seen by large negative differences. \uline{Category} corresponds with attributes determined to be object-level.}
    \label{fig:stylegan-result-bias}
\end{figure}

We present analyses in Fig.~\ref{fig:gpt-sweep-results} for the three real datasets with rich attribute labels: DeepFashion, CelebA, and CUB-200. We swept $N$ in $\{3, 5, 10, 15, 20\}$ and $M$ in $\{3, 5, 7, 10, 15, 20\}$. As expected, increasing $N$ and $M$ increases recall and decreases the fraction of effective attributes. For DeepFashion and CelebA with $N \geq 10$, increasing $M$ leads to minimal gains to recall and large dropoffs in the fraction of effective attributes, indicating that GPT starts to generate redundant attributes. For CUB-200, increasing both $N$ and $M$ results in large increases in recall. However, for large $N$, increases in $M$ result in sharper decreases in the fraction of effective attributes. For all datasets, $N=10$ categories and $M=5$ attributes provide a good balance between few attributes and concept coverage. We use these values in subsequent experiments.


\subsection{Analysis of attribute annotation (VLMs)}
\label{subsec:eval-vlms}



Next, we evaluate the performance of BLIP and OWLv2 for annotation. We use the same three real datasets from the previous section (DeepFashion, CelebA, and CUB-200) because they each have ground-truth annotations for many attributes. We predict labels within each category in a multilabel classification scheme. Table~\ref{tab:vlm-results} presents the average AUC across all dataset attributes for each VLM. BLIP shows good performance (AUC $>0.7$) across all datasets and generally outperforms OWLv2 likely due to the larger fraction of image-level attributes labeled in the datasets.

We also evaluated BLIP's performance on image-level attributes using human annotators. We selected one GPT-generated category (not already labeled) from each of four datasets: ``style'' for DeepFashion, ``habitat'' for CUB-200, ``location" for Stanford Cars, and ``color scheme" for SD Living Rooms. We collected human annotations on a subset of each dataset (200-400 images per dataset, 1-3 annotations per image) using 12 unique annotators. Fig.~\ref{fig:human-eval} presents confusion matrices reporting the agreement level between human and VLM annotations. There is good agreement between humans and BLIP on background scenery (i.e., bird habitat and car location). However, for DeepFashion there is confusion between pairs of styles (``bohemian'' vs. ``vintage'' and ``formal'' vs. ``minimalist'') and for color schemes in SD Living Rooms. We provide visual examples of disagreements between humans and VLMs in Supplementary, as well as inter-annotator disagreements.  

\subsection{Discovering biases in real datasets}
\label{subsec:exp-real-datasets}

We next demonstrate using \name{} to reveal biases in real datasets: DeepFashion, CUB-200, and Stanford Cars. 
We show a few example attributes with annotations in the top row of Fig.~\ref{fig:qualitative-examples}. We provide full lists of generated categories and attributes, example annotations, and histograms in Supplementary. Table~\ref{tab:table-bias} presents a summary of the generated categories and their associated attributes with the highest frequencies. There are several attributes our method reports as having high occurrences that are not labeled in the original datasets. For example, over 35\% of images in DeepFashion contain minimalist-style clothing items, over 35\% of images in CUB-200 contain a bird perching on a tree branch, and over 60\% of images in Stanford cars contain a brand-new car. There are generally many uneven attribute distributions within categories.

We also demonstrate using \name{} to reveal confounding biases in these datasets, i.e., correlations between attribute pairs. We combined the existing classification labels from CUB-200 and Stanford Cars with annotations of the bird habitats and car locations generated by \name{}, resulting in an analysis presented in Fig.~\ref{fig:background-bias}. For CUB-200, waterbirds (a bird species known to live on or around water) appear more often in this dataset with water-related environments (e.g. ``coastal" or ``wetland" habitats), consistent with observations from prior work~\cite{sagawa2020distributionally}. In Stanford Cars, luxury brands (costing more than $\$70K$) such as Bugatti or Ferrari, appear less often in parking lots or gas stations.  





\subsection{Discovering biases of generative image models}
\label{subsec:exp-synthetic-datasets}

We next demonstrate using \name{} to evaluate biases of image generation models using the synthetic datasets SD Living Rooms, SG2 Faces, and SG2 Dogs. 
We show example attributes and annotations in the bottom row of Fig.~\ref{fig:qualitative-examples}. We provide a full list of generated categories, attributes, and example annotations in Supplementary. 

We plot a histogram of generated attributes for SD Living Rooms in Fig.~\ref{fig:living-rooms-dist}. Several categories have uneven attribute distributions. For example, over 90\% of generated living rooms contain a coffee table, sofa, area rug, or throw pillows. Furthermore, less than 10\% contain wall sconces, bookshelves, blinds, shutters, or shades. The majority of living rooms also have an ``eclectic'' layout, a ``neutral'' color scheme,'' a ``Bohemian'' or ``Scandinavian'' style, and a ``cozy and rustic'' ambiance. BLIP struggles to annotate generated flooring attributes, with the majority of images receiving a higher score for the base caption. 

Next, we analyze differences in attribute distributions between StyleGAN2 generators and their training distributions (FFHQ and AFHQ-Dogs datasets). We show the differences in attribute frequencies computed by \name{} in Fig.~\ref{fig:stylegan-result-bias}. The analysis demonstrates SG2 amplifies bias --for both SG2 Faces and SG2 Dogs, the majority attribute per category in the training dataset almost always has an exacerbated majority in the generated dataset. This is shown in the plot as a negative difference (i.e. higher frequency in the generated dataset) for several of the first attributes in each category (the attributes are sorted in order of descending frequency in the training dataset). For example, in SG2 Faces, over 10\% more images contain a smiling facial expression, fair skin tone, brunette hair color, middle-aged appearance, hazel eye color, and stubble facial hair in comparison to its corresponding training dataset. For SG2 Dogs, over 20\% more images contain a dog with an ``alert" posture and over 10\% more contain a medium-sized dog in comparison to its training dataset. 
\section{Discussion and Conclusion}
\label{sec:discussion}
We propose \name{}, the first semi-automated framework leveraging the power of large language and vision-language models to suggest and annotate attributes for dataset bias analysis. Experimental results demonstrate that GPT can successfully suggest most attributes already labeled in real datasets, while also suggesting new ones that lead to bias discoveries. The VLMs (BLIP and OWLv2) also perform well, though BLIP struggled with certain image-level attributes like styles (e.g., ``Bohemian'') and color schemes (e.g., ``triadic''). However, as shown in Supplementary, these attributes also have high levels of inter-annotator disagreement, meaning they are difficult even for humans to judge. These findings lead to the insight that GPT should not just be evaluated in isolation in terms of generated attribute coverage, but also in terms of how well its attributes may be confidently labeled without confusion. A future direction is to develop methods to constrain GPT to do so. 

Evaluation of image generation algorithms, particularly large text-to-image models, is drawing interest in the vision community. Given that a model like Stable Diffusion can generate any image distribution describable by text, it is desirable to also develop analysis algorithms like \name{} that are equally flexible. Results demonstrate that Stable Diffusion can skew color schemes, accessories, and furniture when generating ``a photo of a living room.'' Such insight can help practitioners engineer their prompts to steer away from unwanted biased attributes. Results also demonstrate that \name{} can measure bias amplifications of a generator with respect to its training distribution, such as with StyleGAN2-produced faces and dogs. 

\name{} has several limitations. First, it is only as good as its constituent LLM and VLMs, which have their own systematic errors and biases. While VLMs have improved tremendously in the past several years, they are still far from perfect on high-level semantics beyond object recognition. In addition, GPT fails to recall a number of attributes annotated in the real datasets. The combination of these errors indicates that a method like \name{} cannot simply replace humans in an annotation pipeline in terms of attribute coverage or annotation accuracy. Instead, \name{} will be most useful as a fast, flexible, and automated tool to perform coarse dataset analysis, complementing existing annotations. 
Second, our current implementation selects one image-level attribute per category for an image (multiclass classification), though an image can contain multiple attributes together (e.g. clothing items can both be formal and minimalist, or living rooms can have both monochromatic and neutral color schemes). Third, we evaluated \name{} on datasets with ``contained'' domains focusing on one type of scene/object. Datasets with complex natural scenes like MS-COCO~\cite{lin2014microsoft} would pose challenges in attribute generation (a compact prompt cannot describe arbitrary natural scenes) and image-level attribute annotations (object-level annotations should be relatively unharmed). 


\subsection{Ethics and responsible use}
\name{} inherits the biases of its LLM/VLM models, which are themselves trained on potentially biased data distributions. Biases of the LLM will mainly result in missed attribute categories which, while undesirable, are not as problematic as VLM biases. VLM biases can result in incorrect annotations, thereby skewing dataset analyses. These inaccuracies may be particularly harmful when dealing with human-centered datasets like faces for which these models are not tuned for. A user should therefore always exercise caution and visually inspect image annotation results to confirm reasonable labels and understand the limitations of the VLMs. We recommend using \name{} not as a replacement to human perceptual ground truth, but as an efficient, flexible, and low-cost method to complement human annotation in dataset bias benchmarking.

{
    \small
    \bibliographystyle{ieeenat_fullname}
    \bibliography{main}
}

\clearpage
\maketitlesupplementary
\appendix

\renewcommand{\thefigure}{S\arabic{figure}}
\setcounter{figure}{0}

\renewcommand{\thetable}{S\arabic{table}}
\setcounter{table}{0}

\section{Attribute generations with GPT}

We access GPT using the OpenAI Python API\footnote{\url{https://github.com/openai/openai-python}}. For all experiments, we use model version \texttt{gpt-3.5-turbo-1106} and use temperature $\tau=0.3$. For category and example attribute generations using prompt queries $Q1$ and $Q2$, we append the queries with the requests ``Output the categories in one Python list" and ``Output the examples in one Python list" respectively. We heuristically find that this produces generations with consistent output templates that can be parsed automatically. 

Table~\ref{tab:prompt-templates} contains a list of prompt templates used to describe noun-attribute relationship phrases. Table~\ref{tab:gen-attributes-1} and~\ref{tab:gen-attributes-2} contain summarized lists of all generated categories, attributes, and image- or object-level decisions, as well as prompt templates used for experiments in Sec. 4.3 and Sec. 4.4 respectively. 

\section{Extended results for VLM performance} 

\subsection{Evaluation of VLMs on real dataset annotations}
We show extended results for BLIP and OWLv2 attribute annotation performance on the CelebA, DeepFashion, and CUB-200 datasets. Table~\ref{tab:vlm-results-deepfash-celeba} details a full account of the AUC score for each attribute in CelebA and DeepFashion, and Table~\ref{tab:vlm-results-cub} details the average AUC score for each attribute category in CUB-200. 

\subsection{Human evaluation of \name{} annotations}
We recruit 12 unique annotators from our work environment and collect annotations using Labelbox Annotate\footnote{\url{https://labelbox.com/product/annotate/}}. We curate a subset of images from the DeepFashion, CUB-200, Stanford Cars, and SD Living Rooms datasets. For each image, annotators are asked to select all attributes that best describe the image with respect to the specified category (e.g., ``Bohemian" for style in DeepFashion). Annotators are allowed to select more than one attribute, and we provide an ``unknown" label for instances in which the annotator believes no attribute adequately describes the image. The final human annotation for an image is determined using the consensus (majority) attribute. For images where the consensus attribute is the ``unknown" label, we select the attribute with the next highest annotation count if available (else we leave as ``unknown").  

\begin{figure*}[h!]
    \centering
    \includegraphics[width=1\linewidth]{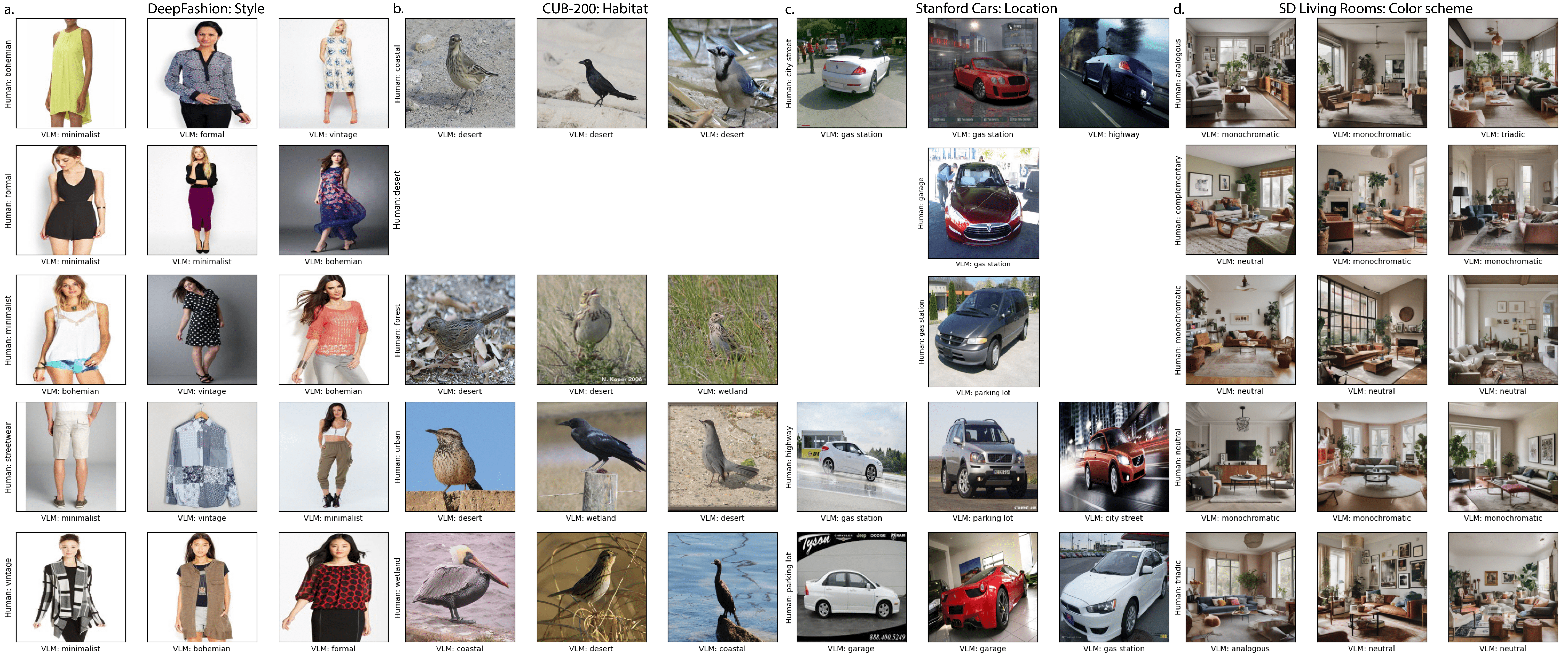}
    \caption{\textbf{Visual samples of disagreement between human and BLIP (VLM) annotations.} Each row per grid corresponds to three randomly selected images determined by a human to contain an attribute that does not match the BLIP annotated attribute. Rows with less than three images correspond with all images in the dataset that have an annotated attribute disagreement (i.e. rows with no images correspond with no disagreements). The corresponding human annotation is given to the left of each row and BLIP annotation below each image. Attribute names have been shortened for brevity.}
    \label{fig:human-vlm-disagreements}
\end{figure*}

Table~\ref{tab:human-annon-breakdown} contains a summary of the human annotations collected for all the datasets used. We observe that there is a higher percentage of images with no consensus attribute obtained for DeepFashion style and SD Living Rooms color scheme, demonstrating the difficulty for humans to label these attributes. Fig.~\ref{fig:human-vlm-disagreements} visualizes examples of images where there is a disagreement between human and BLIP annotations. 

\section{Extended results for \name{}}

\begin{figure*}[h!]
    \centering
    \includegraphics[width=0.925\linewidth]{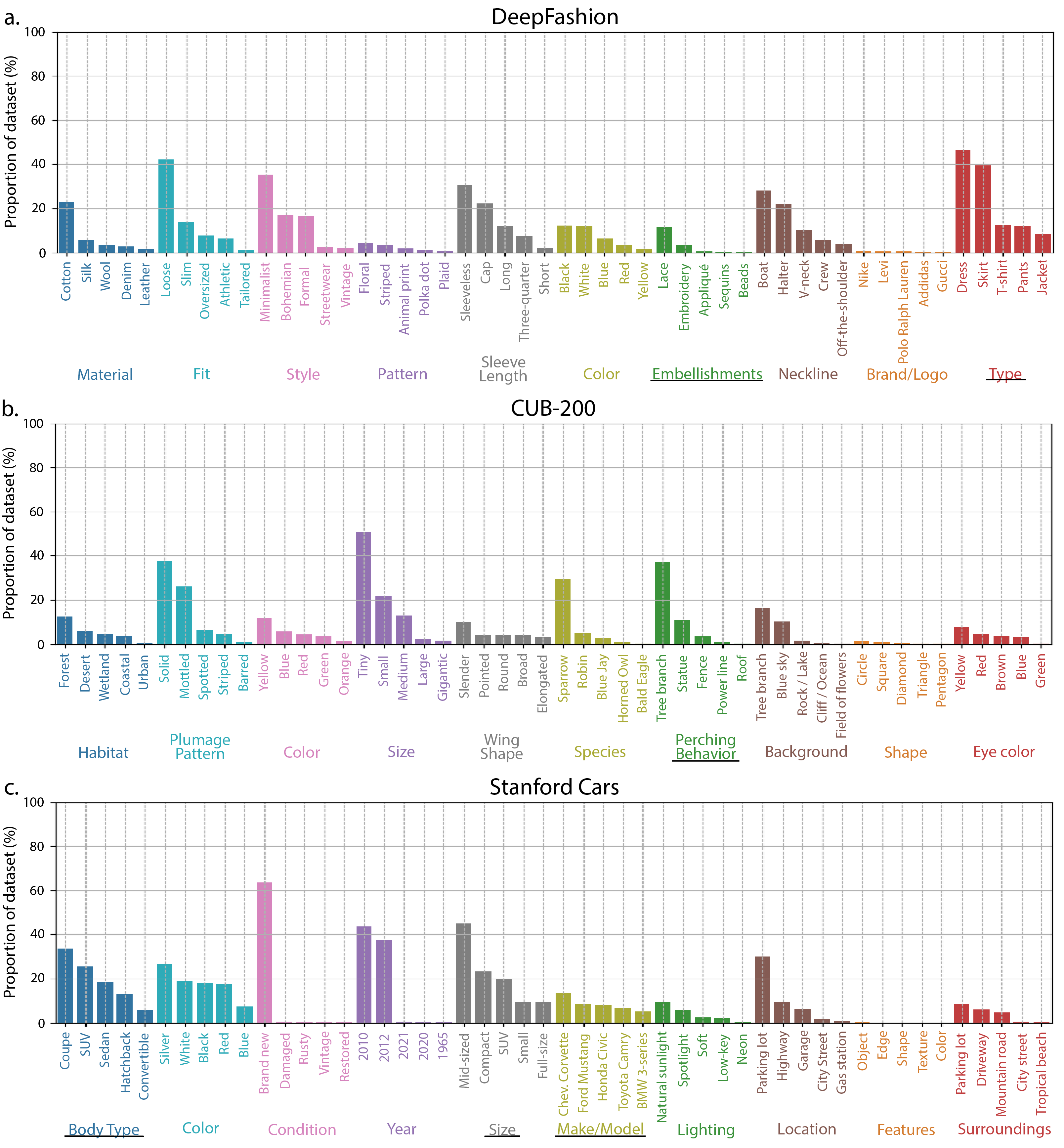}
    \caption{\textbf{Distribution of annotated attributes returned by \name{} for real datasets.} (a.) DeepFashion, (b.) CUB-200, (c.) Stanford Cars. Bars are grouped by attribute categories in different colors, and attribute names have been shortened for brevity. \uline{Category} corresponds with attributes determined to be object-level.}
    \label{fig:all-real-dist}
\end{figure*}

\begin{figure*}[]
    \centering
    \includegraphics[width=1\linewidth]{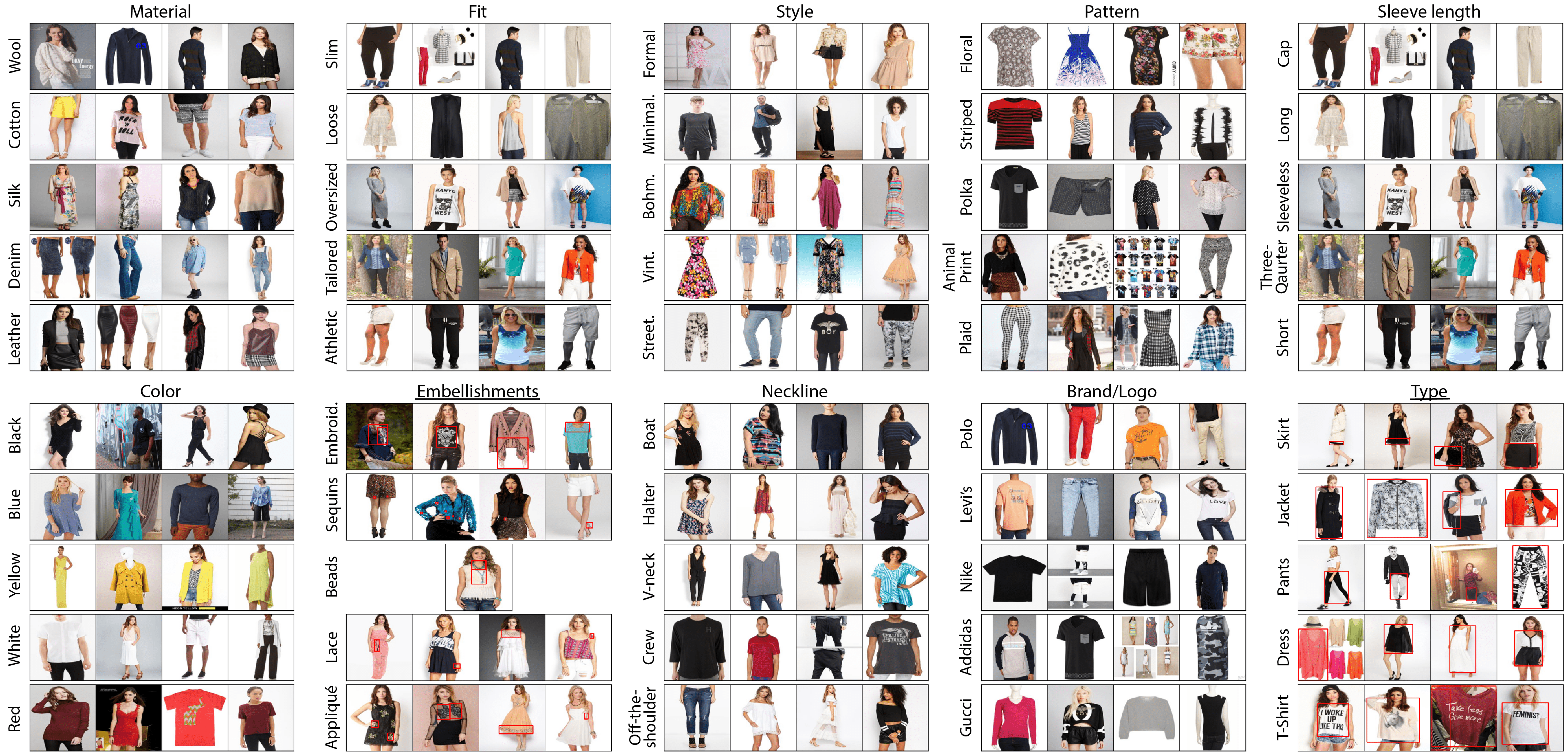}
    \caption{ \textbf{Visual samples with annotations produced by \name{} for DeepFashion.} Each row per grid corresponds to four randomly selected images containing the specified attribute from a category. Rows with less than four images correspond with all images in the dataset determined to contain the attribute. Attribute names have been shortened for brevity. \uline{Category} corresponds with attributes determined to be object-level, and are also shown in images with red bounding box detections.}
    \label{fig:deepfashion-examples}
\end{figure*}

\begin{figure*}[]
    \centering
    \includegraphics[width=1\linewidth]{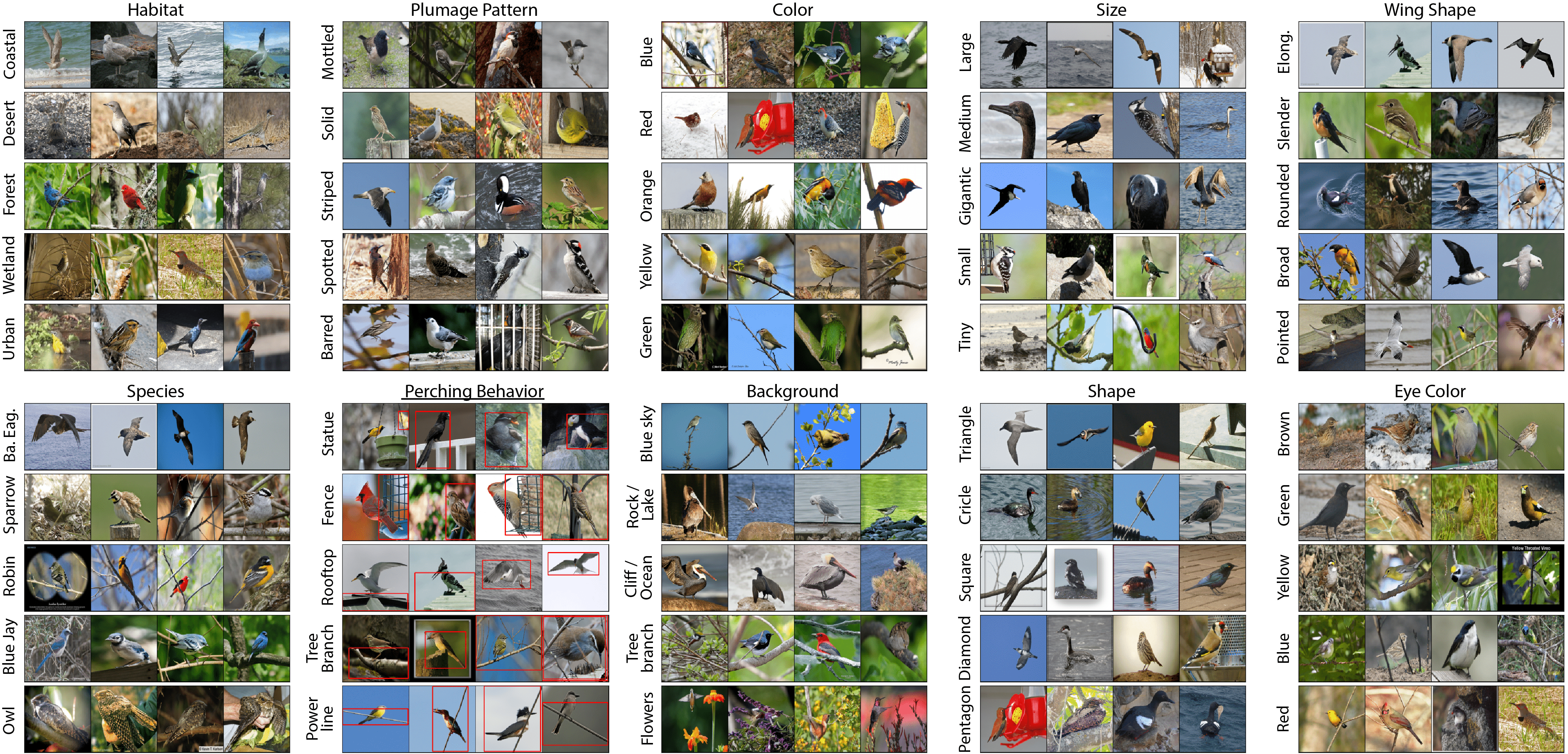}
    \caption{\textbf{Visual samples with annotations produced by \name{} for CUB-200.} Each row per grid corresponds to four randomly selected images containing the specified attribute from a category. Rows with less than four images correspond with all images in the dataset determined to contain the attribute. Attribute names have been shortened for brevity. \uline{Category} corresponds with attributes determined to be object-level, and are also shown in images with red bounding box detections.}
    \label{fig:cub-examples}
\end{figure*}

\begin{figure*}[]
    \centering
    \includegraphics[width=1\linewidth]{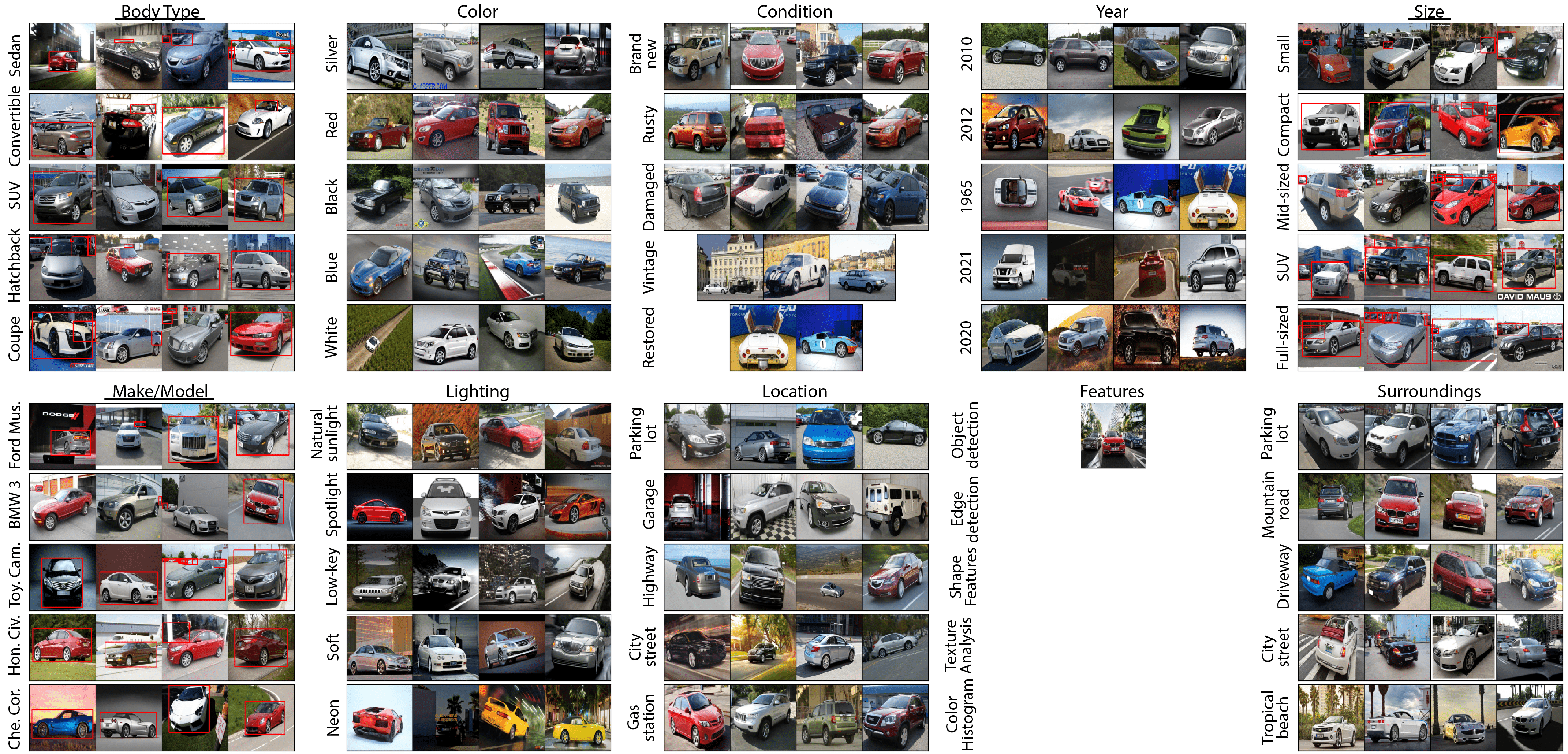}
    \caption{\textbf{Visual samples with annotations produced by \name{} for Stanford Cars.} Each row per grid corresponds to four randomly selected images containing the specified attribute from a category. Rows with less than four images correspond with all images in the dataset determined to contain the attribute (i.e. rows with no images correspond with no annotations). Attribute names have been shortened for brevity. \uline{Category} corresponds with attributes determined to be object-level, and are also shown in images with red bounding box detections.}
    \label{fig:stanford-cars-examples}
\end{figure*}

\begin{figure*}[]
    \centering
    \includegraphics[width=1\linewidth]{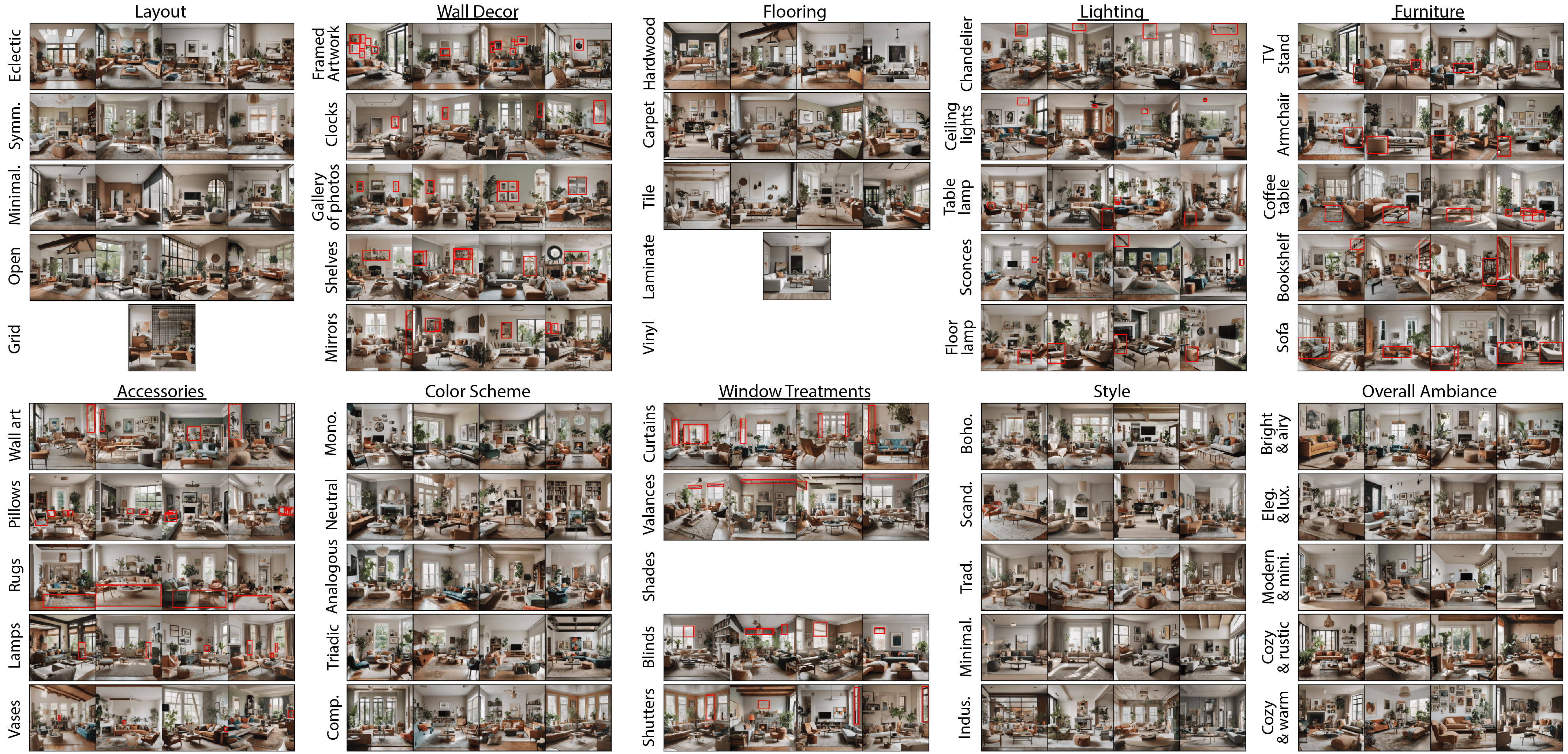}
    \caption{\textbf{Visual samples with annotations produced by \name{} for SD Living Rooms.} Each row per grid corresponds to four randomly selected images containing the specified attribute from a category. Rows with less than four images correspond with all images in the dataset determined to contain the attribute (i.e. rows with no images correspond with no annotations). Attribute names have been shortened for brevity. \uline{Category} corresponds with attributes determined to be object-level, and are also shown in images with red bounding box detections.}
    \label{fig:living-room-examples}
\end{figure*}

\begin{figure*}[]
    \centering
    \includegraphics[width=1\linewidth]{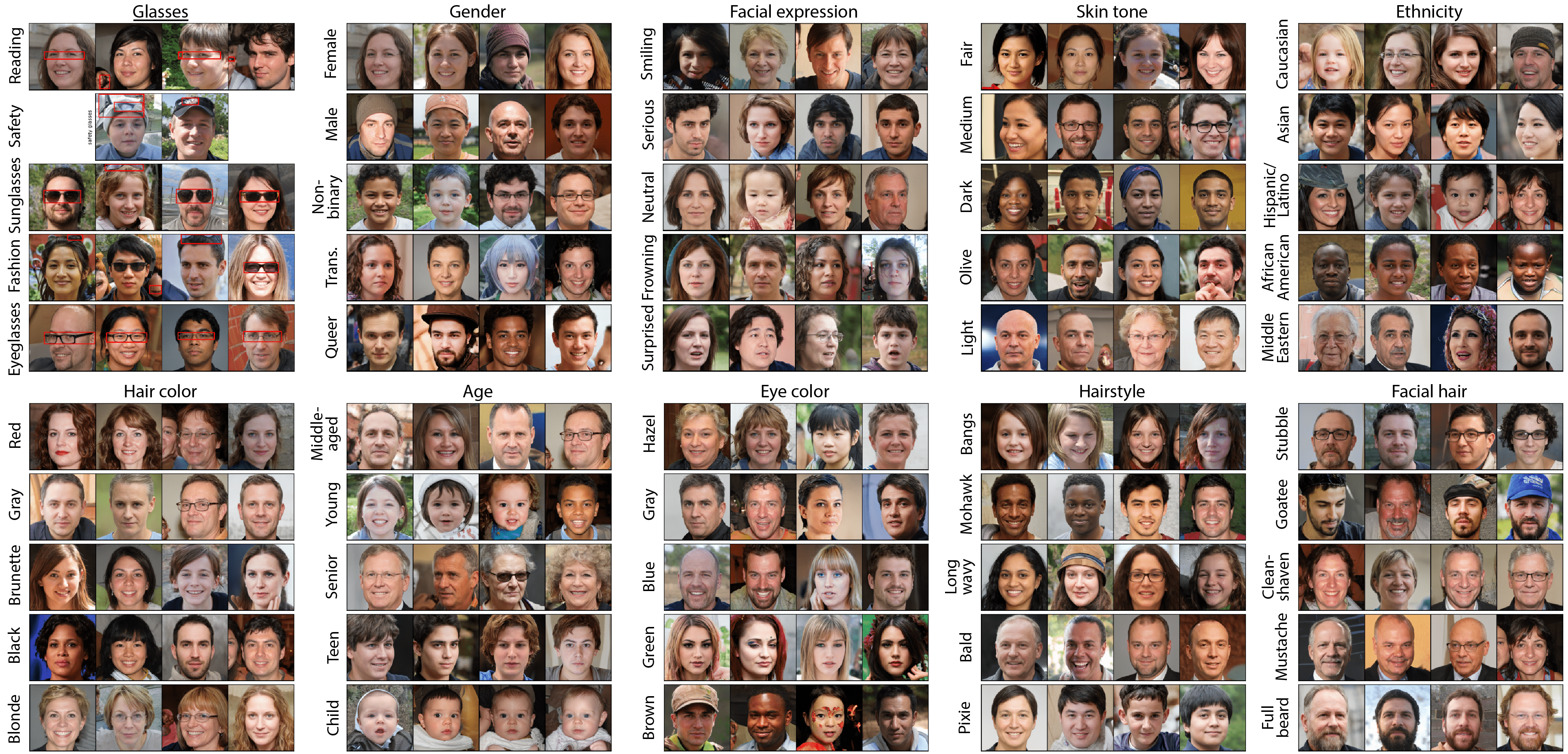}
    \caption{\textbf{Visual samples with annotations produced by \name{} for SG2 Faces.} Each row per grid corresponds to four randomly selected images containing the specified attribute from a category. Rows with less than four images correspond with all images in the dataset determined to contain the attribute. Attribute names have been shortened for brevity. \uline{Category} corresponds with attributes determined to be object-level, and are also shown in images with red bounding box detections.}
    \label{fig:sg2-ffhq-examples}
\end{figure*}

\begin{figure*}[]
    \centering
    \includegraphics[width=1\linewidth]{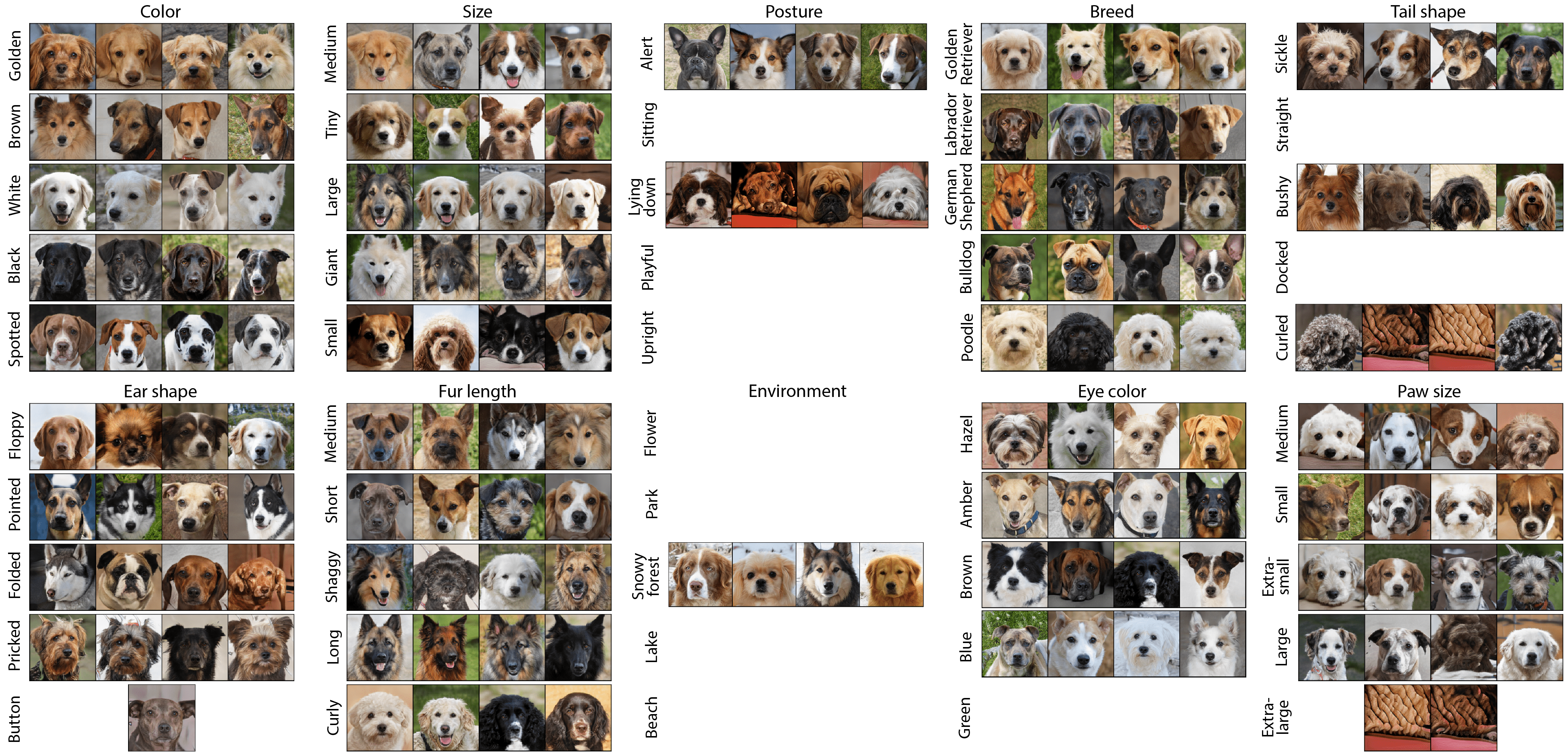}
    \caption{\textbf{Visual samples with annotations produced by \name{} for SG2 Dogs.} Each row per grid corresponds to four randomly selected images containing the specified attribute from a category. Rows with less than four images correspond with all images in the dataset determined to contain the attribute (i.e. rows with no images correspond with no annotations). Attribute names have been shortened for brevity. \uline{Category} corresponds with attributes determined to be object-level, and are also shown in images with red bounding box detections.}
    \label{fig:sg2-dog-examples}
\end{figure*}

Fig.~\ref{fig:all-real-dist} plots histograms of generated attributes for all the datasets used in Sec. 4.3. Fig.~\ref{fig:deepfashion-examples}, \ref{fig:cub-examples}, \ref{fig:stanford-cars-examples}, \ref{fig:living-room-examples}, \ref{fig:sg2-ffhq-examples}, and \ref{fig:sg2-dog-examples} visualize attribute annotations for images across all datasets used in Sec. 4.3 and Sec. 4.4. 

\clearpage
\begin{table}[h!]
\centering
\small
\caption{\textbf{Prompt templates corresponding to a noun-attribute relationship phrase}. Each row details the verb or preposition used to establish the noun-attribute relationship and the corresponding prompt template. \texttt{noun}, \texttt{attr}, and \texttt{category} correspond with the noun describing the image domain, generated attribute, and generated category respectively.}
\label{tab:prompt-templates}
\begin{tabular}{l|l}
\textit{Verb/Preposition} & \textit{Template} \\ \hline
is & a \texttt{attr} \texttt{noun} \\
has & a \texttt{noun} has \texttt{attr} \texttt{category} \\
with & a \texttt{noun} with \texttt{attr} \\
in & a \texttt{noun} in \texttt{attr} \\
from & a \texttt{noun} from \texttt{attr}
\end{tabular}
\end{table}

\begin{table*}[h!]
\centering
\footnotesize
\caption{\textbf{Attributes generated by GPT for real datasets}. We detail the generated attributes used in experiments from Sec. 4.3. The template column corresponds with the prompt template from Table~\ref{tab:prompt-templates} used to convert the attributes into a text caption. In instances where GPT generates attributes that are already in a caption-like form or the attributes are determined to be object-level, we do not use a prompt template. \uline{Category} corresponds with attributes determined to be object-level.} 
\label{tab:gen-attributes-1}
\begin{tabular}{cp{1in}p{0.5in}p{4.3in}}
\toprule
\multicolumn{1}{l}{Dataset} & Category & Template & Attributes \\ \hline 
\multirow{10}{*}[-2em]{\rotcell{\begin{tabular}[c]{@{}c@{}}\textbf{DeepFashion}\\``a photo of a\\clothing item"\end{tabular}}} & Material & is & ``silk", ``leather", ``cotton", ``denim", ``wool"\\
& Fit & with & ``loose fit", ``oversized fit", ``slim fit", ``tailored fit", ``athletic fit"\\
& Style & with & ``bohemian style", ``vintage style", ``streetwear style", ``formal style", ``minimalist style"\\
& Pattern & with &  ``striped pattern", ``floral pattern", ``polka dot pattern", ``animal print pattern", ``plaid pattern"\\
& Sleeve length & with & ``sleeveless", ``short sleeve", ``long sleeve", ``cap sleeve", ``three-quarter sleeve"\\
& Color & is & ``red", ``yellow", ``blue", ``black", ``white"\\
& {\ul Embellishments} & - & ``embroidery", ``sequins", ``beads", ``lace", ``appliqu\'e"\\
& Neckline & with & ``v-neckline", ``halter neckline", ``boat neckline",``off-the-shoulder neckline", ``crew neckline"\\
& Brand/logo & with & ``gucci logo", ``adidas logo", ``nike logo", ``polo ralph lauren logo", ``levi's logo"\\
& {\ul Type of clothing item} & - & ``skirt", ``jacket", ``pants", ``dress", ``t-shirt"  \\ \hline
\multirow{10}{*}[-5.5em]{\rotcell{\begin{tabular}[c]{@{}c@{}}\textbf{CUB-200}\\``a photo of a bird"\end{tabular}}} & Habitat & in & ``forest habitat", ``wetland habitat", ``desert habitat", ``coastal habitat", ``urban habitat"\\
& Plumage pattern & has & ``solid-colored", ``striped", ``spotted", ``mottled", ``barred"\\
& Color & is & ``yellow", ``blue", ``orange", ``green", ``red"\\
& Size & is & ``small", ``medium-sized", ``large", ``tiny", ``gigantic"\\
& Wing shape & has & ``rounded wings", ``pointed wings", ``broad wings", ``slender wings", ``elongated wings"\\
& Species & is & ``american robin", ``great horned owl", ``bald eagle", ``blue jay", ``sparrow" \\
& {\ul Perching behavior} & - & ``bird perched on a statue", ``bird perched on a fence", ``bird perched on a rooftop", ``bird perched on a tree branch", ``bird perched on a power line"\\
& Background scenery & - & ``a photo of a bird with a clear blue sky as the background scenery", ``a photo of a bird perched on a tree branch with lush green foliage as the background scenery", ``a photo of a bird flying above a field of colorful wildflowers as the background scenery", ``a photo of a bird standing on a rock with a serene lake as the background scenery", ``a photo of a bird standing on a rocky cliff with a vast ocean as the background scenery"\\
& Shape & has & ``square", ``circle", ``triangle", ``pentagon", ``diamond"\\
& Eye color & has & ``yellow eyes", ``red eyes", ``blue eyes", ``green eyes", ``brown eyes" \\ \hline
\multirow{10}{*}[-4em]{\rotcell{\begin{tabular}[c]{@{}c@{}}\textbf{Stanford Cars}\\``a photo of a car"\end{tabular}}} & {\ul Body type} & - & ``sedan", ``convertible", ``SUV", ``hatchback", ``coupe"\\
& Color & is & ``blue", ``red", ``black", ``white", ``silver"\\
& Condition & is & ``vintage", ``brand new", ``damaged", ``restored", ``rusty"\\
& Year & from & ``1965", ``2010", ``2021", ``2012", ``2020"\\
& {\ul Size} & - & ``small car", ``compact car", ``mid-size car", ``SUV", ``full-size car"\\
& {\ul Make/model} & - & ``ford mustang", ``bmw 3 series", ``toyota camry", ``honda civic", ``chevrolet corvette"\\
& Lighting & in & ``natural sunlight", ``spotlight", ``soft lighting", ``dramatic low-key lighting", ``neon lighting"\\
& Location & - & ``a photo of a car in a parking lot", ``a photo of a car on a city street", ``a photo of a car in a garage", ``a photo of a car on a highway", ``a photo of a car at a gas station"\\
& Features & - & ``color histogram", ``object detection", ``texture analysis", ``shape features", ``edge detection"\\
& Surroundings & - & ``a photo of a car parked in a busy city street", ``a photo of a car surrounded by palm trees on a tropical beach", ``a photo of a car driving on a winding mountain road", ``a photo of a car in a crowded parking lot", ``a photo of a car parked in a suburban driveway"
\end{tabular}
\end{table*}

\begin{table*}
\centering
\footnotesize
\caption{\textbf{Attributes generated by GPT for synthetic datasets}. We detail the generated attributes used in experiments from Sec. 4.4. The template column corresponds with the prompt template from Table~\ref{tab:prompt-templates} used to convert the attributes into a text caption. In instances where GPT generates attributes that are already in a caption-like form or the attributes are determined to be object-level, we do not use a prompt template. \uline{Category} corresponds with attributes determined to be object-level.} 
\label{tab:gen-attributes-2}
\begin{tabular}{cp{1in}p{0.5in}p{4.3in}}
\toprule
\multicolumn{1}{l}{Dataset} & Category & Template & Attributes \\ \hline 
\multirow{10}{*}[-6.5em]{\rotcell{\begin{tabular}[c]{@{}c@{}}\textbf{SD Living Rooms}\\``a photo of a living room"\end{tabular}}} & Layout & with & ``symmetrical layout", ``eclectic layout", ``grid layout", ``minimalist layout", ``open layout"\\
& {\ul Wall decor} & - & ``framed artwork or paintings", ``wall clocks", ``gallery wall with a collection of framed photos or prints", ``wall-mounted shelves with decorative items or plants", ``wall mirrors"\\
& Flooring & with & ``hardwood flooring", ``laminate flooring", ``carpet flooring", ``tile flooring", ``vinyl flooring"\\
& {\ul Lighting} & - & ``chandelier hanging from the ceiling", ``recessed ceiling lights", ``table lamp on a side table", ``wall sconces on either side of the fireplace", ``floor lamp next to a cozy armchair"\\
& {\ul Furniture} & - & ``tv stand", ``armchair", ``coffee table", ``bookshelf", ``sofa"\\
& {\ul Accessories} & - & ``wall art", ``throw pillows", ``area rugs", ``table lamps", ``decorative vases"\\
& Color scheme & with & ``complementary color scheme", ``monochromatic color scheme", ``analogous color scheme", ``triadic color scheme", ``neutral color scheme"\\
& {\ul Window treatments} & - & ``curtains", ``valances", ``shades", ``blinds", ``shutters"\\
& Style & with & ``industrial style", ``bohemian style", ``traditional style", ``scandinavian style", ``minimalist style" \\
& Overall ambiance & is & ``elegant and luxurious", ``modern and minimalist", ``bright and airy", ``cozy and rustic", ``cozy and warm" \\ \hline
\multirow{10}{*}[-3em]{\rotcell{\begin{tabular}[c]{@{}c@{}}\textbf{SG2 Faces}\\``a headshot photo\\of a person"\end{tabular}}} & {\ul Glasses} & - & ``reading glasses", ``safety glasses", ``sunglasses", ``fashion glasses", ``eyeglasses"\\
& Gender & is & ``genderqueer", ``female", ``non-binary", ``male", ``transgender" \\
& Facial expression & has & ``neutral", ``smiling", ``frowning", ``serious", ``surprised"\\
& Skin tone & with & ``medium skin tone", ``dark skin tone", ``olive skin tone", ``light skin tone", ``fair skin tone"\\
& Ethnicity & is & ``african American", ``caucasian", ``asian", ``middle eastern", ``hispanic/latino" \\
& Hair color & with & ``blonde hair", ``brunette hair", ``red hair", ``gray hair", ``black hair" \\
& Age & is & ``middle-aged", ``child", ``teenager", ``young", ``senior"\\
& Eye color & has & ``blue eyes", ``hazel eyes", ``green eyes", ``gray eyes", ``brown eyes"\\
& Hairstyle & with & ``pixie cut", ``bald head, ``mohawk", ``bangs", ``long wavy hair"\\
& Facial hair & with & ``clean-shaven", ``goatee", ``full beard", ``stubble", ``mustache" \\ \hline
\multirow{10}{*}[-5em]{\rotcell{\begin{tabular}[c]{@{}c@{}}\textbf{SG2 Dogs}\\``a photo of a dog"\end{tabular}}} & Color & is & ``black", ``white", ``golden", ``brown", ``spotted"\\
& Size & is & ``giant", ``tiny", ``medium-sized", ``small", ``large"\\
& Posture & with & ``upright posture", ``playful posture", ``sitting posture", ``alert posture", ``lying down posture"\\
& Breed & is & ``labrador retriever", ``poodle", ``golden retriever", ``bulldog", ``german shepherd"\\
& Tail shape & has & ``straight tail", ``bushy tail", ``docked tail", ``sickle-shaped tail", ``curled tail"\\
& Ear shape & with & ``floppy ears", ``folded ears", ``pointed ears", ``pricked ears", ``button ears"\\
& Fur length & with & ``long fur", ``short fur", ``shaggy fur", ``medium fur", ``curly fur"\\
& Environment & - & ``a photo of a dog playing in a lush green park", ``a photo of a dog walking on a sandy beach with crashing waves in the background", ``a photo of a dog swimming in a crystal-clear lake", ``a photo of a dog sitting in a snowy forest during winter", ``a photo of a dog exploring a colorful flower garden"\\
& Eye color & has & ``green eyes", ``hazel eyes", ``amber eyes", ``brown eyes", ``blue eyes"\\
& Paw size & with & ``large paw size", ``extra-large paw size", ``small paw size", ``medium paw size", ``extra-small paw size"
\end{tabular}
\end{table*}


\begin{table}
\centering
\scriptsize
\caption{\textbf{VLM performances on ground-truth attribute labels from DeepFashion and CelebA.} Numbers are AUC across all attributes in the dataset. We observe good performance ($>0.7$) of BLIP and OWLv2 across the majority of attributes.}
\label{tab:vlm-results-deepfash-celeba}
\begin{tabular}{llcc}
\toprule
Dataset & Attribute & BLIP & OWLv2 \\ \hline
\multirow{24}{*}{DeepFashion} & Floral & 0.80 & 0.84 \\
 & Graphic & 0.76 & 0.61 \\
 & Striped & 0.94 & 0.90 \\
 & Embroidered & 0.69 & 0.55 \\
 & Pleated & 0.71 & 0.65 \\
 & Solid & 0.69 & 0.35 \\
 & Lattice & 0.64 & 0.68 \\
 & Long sleeve & 0.91 & 0.96 \\
 & Short sleeve & 0.92 & 0.93 \\
 & Sleeveless & 0.87 & 0.67 \\
 & Maxi length & 0.94 & 0.82 \\
 & Mini length & 0.81 & 0.49 \\
 & No dress & 0.48 & 0.24 \\
 & Crew neckline & 0.70 & 0.58 \\
 & V-neckline & 0.66 & 0.61 \\
 & Square neckline & 0.86 & 0.44 \\
 & No neckline & 0.60 & 0.46 \\
 & Denim & 0.93 & 0.83 \\
 & Chiffon & 0.86 & 0.74 \\
 & Cotton & 0.59 & 0.44 \\
 & Leather & 0.92 & 0.89 \\
 & Faux & 0.93 & 0.74 \\
 & Knit & 0.87 & 0.83 \\
 & Tight & 0.82 & 0.43 \\
 & Loose & 0.75 & 0.54 \\
 & Conventional & 0.53 & 0.58 \\ \hline
\multirow{40}{*}{CelebA} & 5 o'clock shadow & 0.33 & 0.87 \\
 & Arched eyebrows & 0.61 & 0.68 \\
 & Attractive & 0.77 & 0.42 \\
 & Bags under eyes & 0.34 & 0.36 \\
 & Bald & 0.97 & 0.96 \\
 & Bangs & 0.90 & 0.79 \\
 & Bip lips & 0.60 & 0.61 \\
 & Big nose & 0.43 & 0.53 \\
 & Black hair & 0.92 & 0.69 \\
 & Blond hair & 0.96 & 0.91 \\
 & Blurry & 0.73 & 0.62 \\
 & Brown hair & 0.83 & 0.69 \\
 & Bushy eyebrows & 0.71 & 0.70 \\
 & Chubby & 0.60 & 0.61 \\
 & Double chin & 0.66 & 0.80 \\
 & Eyeglasses & 0.99 & 0.99 \\
 & Goatee & 0.90 & 0.97 \\
 & Gray hair & 0.94 & 0.90 \\
 & Heavy makeup & 0.83 & 0.75 \\
 & High cheekbones & 0.25 & 0.44 \\
 & Male & 0.99 & 0.97 \\
 & Mouth slightly open & 0.56 & 0.57 \\
 & Mustache & 0.80 & 0.92 \\
 & Narrow eyes & 0.42 & 0.32 \\
 & Beard & 0.72 & 0.95 \\
 & Oval face & 0.58 & 0.52 \\
 & Pale skin & 0.74 & 0.58 \\
 & Pointy nose & 0.47 & 0.42 \\
 & Receding hairline & 0.62 & 0.67 \\
 & Rosy cheeks & 0.68 & 0.51 \\
 & Sideburns & 0.89 & 0.83 \\
 & Smiling & 0.96 & 0.54 \\
 & Straight hair & 0.70 & 0.45 \\
 & Wavy hair & 0.91 & 0.78 \\
 & Wearing earrings & 0.91 & 0.91 \\
 & Wearing hat & 0.97 & 0.96 \\
 & Wearing lipstick & 0.90 & 0.92 \\
 & Wearing necklace & 0.81 & 0.75 \\
 & Wearing necktie & 0.90 & 0.94 \\
 & Young & 0.88 & 0.39
\end{tabular}
\end{table}

\begin{table}[]
\centering
\small
\caption{\textbf{VLM performances on ground-truth attribute labels from CUB-200.} Numbers are average AUC scores over attributes from each category. In general, we observe good performance ($>0.7$) of BLIP and OWLv2 for color categories. Furthermore, performance on attributes localized to a single bird part (e.g., wing or tail) is significantly worse.}
\label{tab:vlm-results-cub}
\begin{tabular}{lcc}
\toprule
Attribute & BLIP & OWLv2 \\ \hline
Primary color & 0.79 & 0.71\\
Upperparts color & 0.74 & 0.72\\
Underparts color & 0.78 & 0.76 \\
Upper tail color &  0.71 & 0.66 \\
Breast color & 0.78 & 0.69\\
Belly color &  0.77 & 0.71\\
Back color &  0.77 & 0.70\\
Bill color & 0.65 & 0.65\\
Leg color &  0.63 & 0.67\\
Forehead color & 0.77 & 0.71\\
Wing color & 0.73 & 0.66\\
Eye color &  0.58 & 0.64\\
Nape color & 0.78 & 0.68\\
Crown color &  0.79 & 0.74\\
Under tail color & 0.72 & 0.64\\
Throat color & 0.76 & 0.67\\
Head pattern &  0.58 & 0.56\\
Breast pattern &  0.59 & 0.57 \\
Belly pattern &  0.57 & 0.58\\
Back pattern & 0.60 & 0.58\\
Wing pattern & 0.60 & 0.55\\
Tail pattern &  0.58 & 0.55 \\
Shape & 0.65 & 0.68\\
Size &  0.68 & 0.68 \\
Bill length & 0.55 & 0.55\\
Bill shape & 0.60 & 0.57 \\
Wing shape & 0.54 & 0.55 \\
Tail shape &  0.55 & 0.53
\end{tabular}
\end{table}

\begin{table}
\small
\centering
\caption{\textbf{Human annotation experiment statistics}. $N_{images}$ corresponds to the total number of images annotated from the dataset. $\overline{N}_{annotations}$ corresponds to the average number of attribute annotations collected per image in the dataset. ``Unknown" corresponds to the percentage of images in which at least one annotator assigned the ``unknown" label to an image. ``No consensus" corresponds to the percentage of images that do not have a consensus attribute.}
\begin{tabular}{l|cccc}
\diagbox{~~}{Dataset\\(Attribute)} & \begin{tabular}[c]{@{}c@{}}DeepFashion\\(Style)\end{tabular} & \begin{tabular}[c]{@{}c@{}}CUB-200\\(Habitat)\end{tabular} & \begin{tabular}[c]{@{}c@{}}Stanford Cars\\(Location)\end{tabular} & \begin{tabular}[c]{@{}c@{}}SD Living Rooms\\(Color scheme)\end{tabular} \\ \hline
$N_{images}$ & 250 & 231 & 250 & 348 \\
$\overline{N}_{annotations}$ & 2.1 & 3.0 & 2.0 & 1.8 \\
Unknown & 4.8\% & 23.4\% & 10.0\% & 0.3\% \\
No consensus & 33.6\% & 16.5\% & 12.6\% & 28.4\%
\end{tabular}
\label{tab:human-annon-breakdown}
\end{table}

\end{document}